\documentclass[11pt,a4paper]{article}
\usepackage{times,latexsym}
\usepackage{url}
\usepackage[T1]{fontenc}

\usepackage[acceptedWithA]{tacl2021v1}
\usepackage{tacl2021v1}

\usepackage{xspace,mfirstuc,tabulary}

\newif\iftaclinstructions
\taclinstructionsfalse 
\iftaclinstructions
\renewcommand{\confidential}{}
\renewcommand{\anonsubtext}{(No author info supplied here, for consistency with
TACL-submission anonymization requirements)}
\newcommand{\instr}
\fi

\iftaclpubformat 

\else

\fi

\usepackage{graphicx,times}
\usepackage{subfigure}         
\usepackage{natbib}
\usepackage{svg}
\usepackage{amssymb,amsmath}
\usepackage{makecell}

\usepackage{amsmath}
\usepackage{booktabs}
\usepackage{multirow}
\usepackage{dsfont}
\usepackage{bbm}
\usepackage{bm}
\usepackage{microtype}
\usepackage{xcolor}
\usepackage{graphicx}
\usepackage[utf8]{inputenc}
\usepackage{bbding}
\usepackage{xcolor}
\usepackage{hyperref}
\newcommand{\ours}{\textsc{MACSum}}

\usepackage{pifont}
\newcommand{\tick}{\ding{51}}
\newcommand{\cross}{\ding{55}}

\definecolor{amber}{rgb}{1.0, 0.75, 0.0}

\title{\ours: Controllable Summarization with Mixed Attributes}

\author{
  Yusen Zhang$^1$\Thanks{\noindent Yusen Zhang completed this work during his internship at Microsoft.} \quad
  Yang Liu$^2$\Thanks{\noindent Correspondence author} \quad Ziyi Yang$^2$ \\
  {\bf
  Yuwei Fang$^2$ \quad
  Yulong Chen$^3$ \quad Dragomir Radev$^4$}\\
  {\bf
  Chenguang Zhu$^2$ \quad
  Michael Zeng$^2$ \quad
  Rui Zhang$^1$} \\
  $^1$Penn State University \quad $^2$Microsoft Research \quad $^3$Westlake University \quad 
  $^4$Yale University\\
  \tt{\{yfz5488,rmz5227\}@psu.edu; yaliu10@microsoft.com}
}

\date{}

\begin{document}
\maketitle
\begin{abstract}
Controllable summarization allows users to generate customized summaries with specified attributes. However, due to the lack of designated annotations of controlled summaries, existing works have to craft pseudo datasets by adapting generic summarization benchmarks. Furthermore, most research focuses on controlling single attributes individually (e.g., a short summary \textbf{or} a highly abstractive summary) rather than controlling a mix of attributes together (e.g., a short \textbf{and} highly abstractive summary).
In this paper, we propose \ours, the first human-annotated summarization dataset for controlling mixed attributes.
It contains source texts from two domains, news articles and dialogues, with human-annotated summaries controlled by five designed attributes (Length, Extractiveness, Specificity, Topic, and Speaker).
We propose two simple and effective parameter-efficient approaches for the new task of mixed controllable summarization based on hard prompt tuning and soft prefix tuning.
Results and analysis demonstrate that hard prompt models yield the best performance on most metrics and human evaluations. However, mixed-attribute control is still challenging for summarization tasks. Our dataset and code are available at~\url{https://github.com/psunlpgroup/MACSum}. 
\end{abstract}

\section{Introduction}
Text summarization is the task of compressing the input text into a concise and coherent version by preserving salient information. 
There has been substantial progress in generic summarization by generating one overall summary for each input~\cite{mckeown1995generating,erkan2004lexrank,rush2015neural,cheng2016neural,see2017get,paulus2018deep}.
However, readers have diverse preferences when summarizing the same article, such as topics, speakers, or lengths of the summary~\citep{fan2018controllable,zhong-etal-2021-qmsum,goyal2021hydrasum}. Therefore, generating customized summaries to meet different preferences is a natural capability of summarization systems.

Due to the lack of a human-annotated controllable summarization benchmark, existing research has to adapt generic datasets to create pseudo-controllable summaries~\citep{fan2018controllable,he2020ctrlsum,zhong-etal-2021-qmsum,goyal2021hydrasum,chan2021controllable}.
\citet{he2020ctrlsum}, for example, extract topics from a generic summary by assuming the summary is controlled by the extracted topics to evaluate summarization over topics.
However, this adaptation-based setting raises three issues. 
First, the adapted summaries are not really written with the guidance of being controlled by the designed attributes.
Second, this conversion can only build one target summary for each source, while it is preferable to have summaries with different control attributes for the \textit{same} input article.
Third, for attributes like Extractiveness or Specificity, there are no straightforward adaptation methods.

\begin{figure}[t!]
    \centering
    \includegraphics[width=\linewidth]{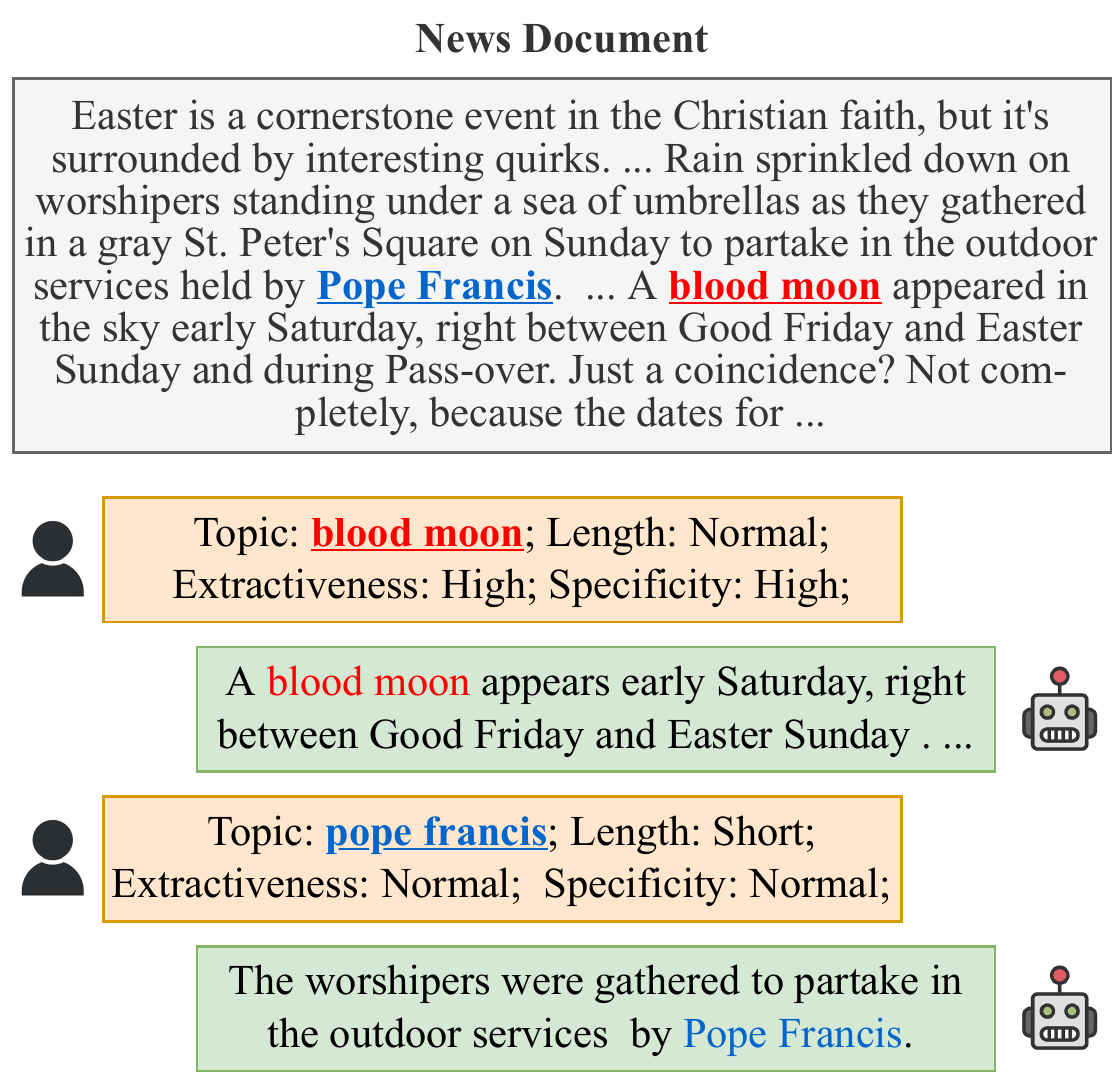}
    \caption{An example of \ours. For the same input source text, the system needs to generate different reference summaries (green boxes) for different mixed control attributes (orange boxes). 
    }
    \label{fig:input}
\end{figure}

Meanwhile, previous studies mostly focus on controlling a single attribute, e.g., generating a short summary \textbf{or} a highly abstractive summary. However, mixing different control attributes is more challenging and underexplored~\citep{russo2020control}. For example, Figure~\ref{fig:input} shows a case of mixed-attribute control by requiring a short summary regarding ``Pope Francis", or a highly extractive and highly specific summary on the topic ``blood moon". Users can simultaneously control different attributes in the generated summary.

In this paper, we propose \ours, a human-annotated benchmark for controllable summarization with mixed attributes. In \ours, source texts are collected from both news and dialogue domains. We define five control attributes of summarization by synthesizing previous studies~\cite{chan2021controllable, liu2018controlling, fan2018controllable}, including Length (\textit{Len}), Extractiveness (\textit{Ext}), Specificity (\textit{Spe}), Topic (\textit{Tpc}), and Speaker (\textit{Spk})\footnote{The speaker attribute is for the dialogue domain only.}.
For each input source, we sample a set of different combinations of these attributes for human annotations.
The resulting \ours~dataset contains a rich set of annotations of human-written summaries for the same input with different mixtures of control attributes. Table~\ref{tab:data_compare} compares \ours~with previous work, and \ours~is the first one to mix these five attributes with human annotations, covering both dialogue and document source texts.

Furthermore, to establish a baseline of mixed-attribute control, we design two simple yet effective frameworks that can steer a summarization model by either hard prompt (HP) or soft prefix (SP) inspired by prompt learning~\citep{raffel2020exploring, li-liang-2021-prefix}. For each value of a control attribute (e.g., long length), in the HP framework, we prepend the description of control attributes (e.g., ``Length: Long'') to the input source as hard prompts; in the SP framework, we assign a set of external parameters, called soft prefixes, to the model. The summarization model is trained to summarize an input with hard prompt/soft prefixes of different control signals. 
We evaluate these baseline models on \ours~ with proposed two automatic evaluation metrics measuring the quality of control. Our results and analysis in two domains reveal that the HP framework yields the best performance on all automatic metrics and human evaluations, however, mixed-attribute control is still challenging.

\begin{table*}[ht!]
\resizebox{\textwidth}{!}{
\begin{tabular}{@{}lccccccccccc@{}}
\toprule
         & \multirow{2}{*}{Domain} & \multicolumn{2}{c}{Source Type} & \multicolumn{2}{c}{Construction} & \multirow{2}{*}{Mixed Attr.} & \multicolumn{5}{c}{Control Attributes} \\ \cmidrule(lr){3-4} 
         \cmidrule(lr){5-6}\cmidrule(lr){8-12}
         &                   &  Dial.  & Doc. & Anno. & Multi-O. & & \textit{Tpc} & \textit{Spk} & \textit{Len} & \textit{Ext} & \textit{Spe}  \\ \midrule
CASum~\citep{fan2018controllable}     & News   & \cross     & \tick        & \cross & \cross & \cross     & \tick     & \tick       & \tick    & \cross    & \cross    \\
CTRLSum~\citep{he2020ctrlsum}  & News, Papers  & \cross     & \tick & \cross &\cross    & \cross     & \tick     & \cross       & \tick    & \cross    & \cross       \\
QMSum~\citep{zhong-etal-2021-qmsum}    & Meetings   & \tick     & \cross             & \tick &\tick & \tick     & \tick     & \tick       & \cross    & \cross    & \cross        \\
HydraSum~\citep{goyal2021hydrasum} & News   & \cross     & \tick                 & \cross &\cross  & \tick     & \cross     & \cross       & \tick    & \tick    & \tick        \\
CMDP~\citep{chan2021controllable}    & News    & \cross     & \tick     & \cross &\cross & \cross     & \tick     & \tick       & \tick    & \tick    & \cross    \\ \midrule
\ours~(ours)     & News, Meetings    & \tick     & \tick      & \tick &\tick & \tick     & \tick     & \tick       & \tick    & \tick    & \tick        \\ \bottomrule
\end{tabular}}
\caption{Comparison between~\ours~and previous work on controllable summarization. Dial. and Doc. means if the source is dialogue or document. Anno. indicates whether the data is constructed by human annotation or rule-based pseudo-split. Multi-O. shows if there are multiple outputs with different control attributes for the same source. Mixed Attr. shows if mixed attribute control is allowed. Control Attributes are defined in Section~\ref{sec:dataset}.
}
\label{tab:data_compare}
\end{table*}

\section{Related Work}

\subsection{Controllable Summarization}
Previous work on controllable text summarization focuses on length~\cite{fan2018controllable,liu2018controlling,makino2019global,saito2020length,liu2022length,he2022z,goyal2022news}, entity~\cite{he2020ctrlsum,narayan2021planning,maddela2022entsum,hofmann2022extractive}, aspect~\cite{tan2020summarizing,amplayo2021aspect}, content~\cite{dou-etal-2021-gsum,https://doi.org/10.48550/arxiv.2212.10819}, style~\cite{cao2021inference}, granularity~\cite{zhong2022unsupervised}, and abstractiveness~\cite{song2020controlling}.
Query-focused summarization~\cite{dang2005overview,fisher2006query,daume2009bayesian} generates summaries for specific user information requests, but it does not explicitly control the output style.
Furthermore, interactive summarization~\cite{bornstein1999interactive,leuski2003ineats} and reinforcement learning guided summarization~\cite{paulus2018deep,bohm2019better,stiennon2020learning} have been used to incorporate human preferences and feedback, yet the human feedback explored so far is largely limited to the generic quality of summaries instead of fine-grained attributes.
Notably, \citet{chan2021controllable} propose a constrained Markov Decision Process for controllable summarization for different attributes, but it is unclear if it can perform multi-attribute control.
\citet{goyal2021hydrasum} investigate multi-feature control by mixing multiple decoders, yet their solution is only based on decoding improvements which yield suboptimal controlling performance.
Therefore, most previous works are over-specialized for controlling particular attributes, while controlling multiple attributes is still underexplored.
Furthermore, existing works are mostly evaluated on pseudo datasets adapted from generic summarization datasets.

\subsection{Prompt Learning}
Prompt learning is first proposed in GPT-3~\cite{DBLP:conf/nips/BrownMRSKDNSSAA20}, where large pretrained language models can perform desired tasks with the guidance of instructions and examples.
Some efforts explore prompt-tuning using natural language by converting original inputs into cloze-style questions and then tuning language models~\cite{shin2020eliciting,schick2021s,DBLP:journals/corr/abs-2202-04824,DBLP:conf/acl/MinLHZ22}.
However, most of them focus on natural language understanding tasks and usually need a careful selection of prompts.
Instead of using human-crafted tokens, other work explores using continuous vectors as prompts~\cite{lester-etal-2021-power,DBLP:conf/naacl/QinE21,liu2021gpt,li-liang-2021-prefix}.
Among them, prefix-tuning is particularly designed for text generation~\cite{li-liang-2021-prefix}.
Prefix-tuning prepends trainable vectors to each layer of language models as prefixes and keeps other parameters frozen during training.
In this work, we propose two methods for mixed attribute controllable summarization based on prompt-tuning and prefix-tuning, respectively.

\section{The \ours~Dataset }
\label{sec:dataset}
To provide a benchmark for controllable summarization, we propose \ours, a high-quality human-annotated mixed-attribute controlled summarization dataset. Inspired by several previous studies on controllable generation~\cite{chan2021controllable,liu2018controlling,fan2018controllable},  \ours~is annotated with 5 types of control attributes, including Topic, Speaker, Length, Extractiveness, and Specificity (Section~\ref{sec:control}). 

As shown in Figure~\ref{fig:input}, these five attributes can be combined together in various designs (Section~\ref{sec:control}). Besides, Topic and Speaker can have multiple values as well, i.e., more than one speaker or topic to focus on. In annotation, we require the corresponding summary to fulfill all attributes together.  

\begin{figure*}[t!]
    \centering
    \includegraphics[width=\textwidth]{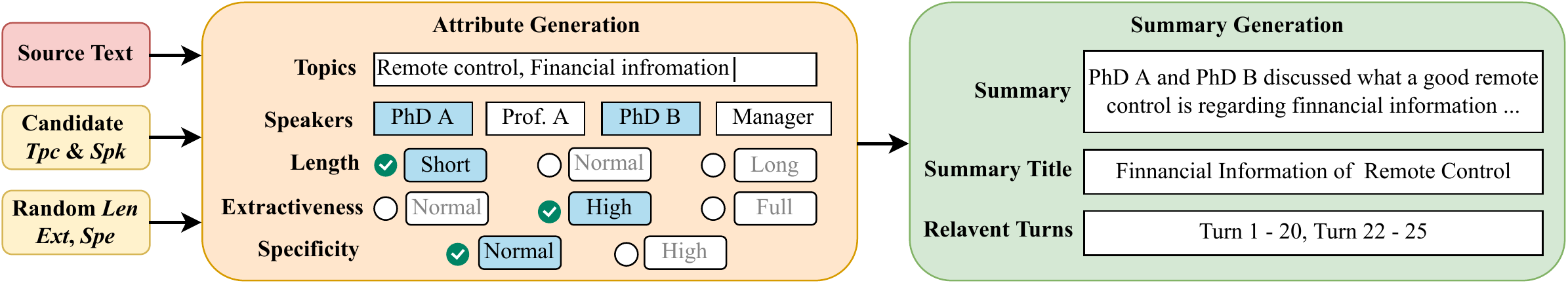}
    \caption{Annotation pipeline of \ours. The annotator needs to summarize the contents of meetings/documents according to the five control attributes, give the relevant text spans, and write a summary title.} 
    \label{fig:annotation}
\end{figure*}

\label{sec:annotation}
The data annotation pipeline is divided into four steps (Figure~\ref{fig:annotation}). First, we carefully select the source texts from several widely used summarization datasets in news or dialogue domains. Second, some automatic tools are leveraged to form a pool of candidate attributes as guidance for the next step. Third, the annotators manually label five attributes to form a control attribute set and repeat the process multiple times. Finally, the annotators write down the summary that meets each control attribute set.

\subsection{Control Attributes}
\label{sec:control}
\ours~provides five attributes to control the summary generation.

\paragraph{Topic (\textit{Tpc})} indicates certain contents of the source that users are particularly interested in. The summary should only contain contents that are related to the given topics. We provide multiple keywords such as ``remote control, financial information'' for annotators as candidate topics.

\paragraph{Speaker (\textit{Spk})} indicates certain speakers in a dialogue whose content is preferred by the user. In \ours~(MAC-Dial only), this is specified by giving the name of certain speakers, such as ``Program Manager''.  

\paragraph{Length (\textit{Len})} indicates the number of words of the summary, serving the time budget for users to read this summary. In \ours, \textit{Len} is controlled by [\textit{short}, \textit{normal}, \textit{long}], three values of this attribute\footnote{We denote attribute value as \textit{Attribute: value}, e.g. \textit{Len: long}.}. Our annotation guideline provides a reference range of compression ratio and word count for each length value.

\paragraph{Extractiveness (\textit{Ext})} describes the proportion of the summary extracted from the source text. This is useful when users sometimes want content directly extracted from the source, while sometimes they want more abstractive and more readable results. In \ours, \textit{Ext} can take values of [\textit{normal}, \textit{high}, \textit{full}].

\paragraph{Specificity (\textit{Spe})} means how many details or descriptive contents we need to include in the summary. Referring to~\citet{louis2011text}, different users can prefer more general summaries or more specific summaries. \ours~contains two levels of \textit{Spe} control, namely [\textit{normal}, \textit{high}], where normal is the natural specificity and high requires more specific contents. 

Specificity differs from Length. Length is the number of words, while Specificity is the density of descriptive contents (e.g., numbers, entities, and names). Thus, a short summary can have a higher Specificity than a long one.

\ours~supports Mixed-Attribute Control because it is a natural need for users to control multiple aspects at the same time, e.g., wanting the summary to be short, highly extractive, and only talking about some topics. To this end, as shown in Figure~\ref{fig:input}, the samples in \ours~can control multiple attributes simultaneously. We require the annotated summaries to meet all requirements at the same time. If some combinations are considered too difficult to fulfill, we allow annotators to skip them in rare cases. We provide detailed distribution of attributes in Figure~\ref{fig:distri}.

\subsection{Annotation Pipeline}

\paragraph{Source Selection}
\ours~covers source text from both document and dialogue summarization tasks. We pick \textbf{CNNDM}~\citep{NIPS2015_afdec700} as the document dataset and \textbf{QMSum}~\citep{zhong-etal-2021-qmsum} as the dialogue dataset. 
CNNDM is a large-scale document summarization dataset containing news stories along with their corresponding highlights, collected from CNN and Daily Mail websites. QMSum is a popular query-based meeting summarization dataset. It contains the transcripts of three domains, including AMI, ICSI, and committee meetings of the Welsh Parliament and the Parliament of Canada.
For CNNDM, we randomly pick 10k documents in the test set for the annotation. 
For QMSum, we first split each meeting into shorter units according to the topic partition and discard the units that are longer than 5000 tokens. 

\paragraph{Attribute Candidate Extraction} For Topic, we first use a keyword extraction tool~\citep{boudin:2016:COLINGDEMO} to extract the top 20 keywords from the source text as candidates. For Speaker, we collect all speakers in the source text to form a candidate set. For the remaining Length, Extractiveness, and Specificity attributes, we generate their values and combination randomly from a uniform distribution, mimicking the behavior of users with diverse needs for customized summaries.

\begin{table*}[!t]
    \centering
    \resizebox{\textwidth}{!}{
    \begin{tabular}{@{}lcccrrrrr@{}}
\toprule
         & \multicolumn{3}{c}{\#Samples/\#Sources} & \multicolumn{3}{c}{Avg. number in Text}     & \multicolumn{2}{c}{Avg. \# C.A.} \\ \cmidrule(lr){2-4} \cmidrule(lr){5-7} \cmidrule(lr){8-9} 
         & Train         & Dev        & Test      & Source Len. & Source Turns & Reference Len. & Topic              & Speaker              \\ \midrule
CNNDM    & 2887k/2887k    & 13k/13k    & 11k/11k   & 781.0         & --             & 56.0             & --                   & --                     \\
QMSum    & 1257/162      & 272/35     & 279/35    & 9069.8      & 556.8        & 69.6           &  --                  &  --                    \\ \midrule
MAC-Doc  & 4278/755      & 554/94     & 547/94    & 835.4      & --             & 54.1          & 0.8               & --                     \\
MAC-Dial & 2338/328      & 292/41     & 324/41    & 2754.3     & 144.6       & 69.4          & 1.7              & 1.2                 \\ \bottomrule
\end{tabular}
    }
    \caption{Statistics of \ours~consisting of two parts: MAC-Doc from CNNDM and MAC-Dial from QMSum. Source Len., Ref. Len. are tokens in source and reference. Topic, Speaker are the averaged number of topics/speakers.}
    \label{tab:statis}
\end{table*}

\paragraph{Attribute Generation} We hire 4 native English speakers as annotators. The annotators can either freely choose topics from the candidate topics or write the keywords by themselves. As for the Speaker attribute, we ask the annotators to pick one or more names from the candidate set. Besides, Length, Extractiveness, and Specificity are automatically filled with randomly generated values. 

Attributes generation repeats several times for each source to form various attribute combinations, so-called samples. Overall, each source text contains eight samples for every two thousand words.

\paragraph{Summary Generation}
\label{sec:annotate}
We first ask all annotators to read our annotation guideline and 10 annotated examples. Afterward, given several combinations of control requirements, i.e., the control attribute sets, the annotators follow our guidance and write a summary for each control combination. 

We also ask them to annotate the related text spans for use in future work, such as retrieval-based methods. Related text spans are the turns/sentences in the source that are most relevant to the golden summary. These spans are the minimum necessary turns/sentences the annotators need to produce the complete summary.

Finally, the annotators read the summary again for quality insurance, and we ask them to write a short title for this summary, e.g., ``discussion of remote control style''. This is helpful for future work such as title generation, and it also provides us with a quick way to verify whether the annotators read their generated summaries.

\paragraph{Quality Control}

First, we control the annotation quality through a careful pilot test. Before the annotation process starts, annotators are carefully selected via a pilot test. We assign each annotator the same three input texts with various mixed attributes, and we choose the qualified annotators according to annotation results. 

Second, we conduct a weekly sampling inspection. We frequently monitor the quality of annotations. We collect the results weekly and provide feedback to the annotators to ensure quality. 

\subsection{Automatic Metrics}
\label{sec:metrics}
\paragraph{Overview}
Along with the annotated benchmark, we also design a system of automatic metrics for evaluating the model's capability to generate controllable summaries.
For each attribute, we define its own attribute metric function to represent the degree of control.
We then propose \textbf{Control Error Rate} (CER) and \textbf{Control Correlation} (CC). 
CER measures the distance between the generated and golden summary in terms of their degrees of control using attribute metric functions. A good model should have smaller CER~$\downarrow$.
CC measures the distribution of attribute metric functions among generated summaries with different attribute values, representing the model's capability to correlate to the definition of the control attribute. A good model should have a CC distribution that is similar to that of the golden summary~$\updownarrow$.
In addition, we also report F-1 of ROUGE-1/2/L~\citep{lin-2004-rouge} for the general quality of the summary~$\uparrow$.

\paragraph{Definition}

For a control attribute $r$ and its attribute metric function $f_r$, given a predicted summary $\hat{y}$, golden summary $y$, Control Error Rate (CER) is defined as:
\begin{equation}
    \text{CER}(\hat{y}, y) = \frac{|f_r(\hat{y}) - f_r(y)|}{f_r(y) + \epsilon} 
    \label{equ:cer}
\end{equation}
where $\epsilon$ is a small value to avoid error when $f_r(y)$ is zero.

Additionally, for the control attribute $r$ (e.g., \textit{Len}) with a control value pair $[v_1, v_2]$ (e.g., [\textit{short}, \textit{long}]), predicted summaries for these two values  $[\hat{y}_1, \hat{y}_2]$, Control Correlation (CC) is defined as:
\begin{equation}
    \text{CC}(\hat{y_1}, \hat{y_2}) = \frac{f_r(\hat{y}_1) - f_r(\hat{y}_2)}{\text{Distance}(v_1, v_2)}
\end{equation}
where $\text{Distance}(v_1, v_2)$ calculates the distance \textit{from} control value $v_1$ \textit{to} $v_2$, which can be negative. For instance, $\text{Distance}(\textit{high}, \textit{normal}) = 1$, and $\text{Distance}(\textit{short}, \textit{long}) = -2$. When CC is above/below 0, it indicates the evaluated model has a positive/negative correlation with the control objective. Additionally, CER and CC for multiple samples are their arithmetic mean.

For each of the five control attributes, we define its own attribute metric $f_r$ which maps the summary to a real number that represents the degree of control. \textbf{Topic} $f_{Tpc}$ is the proportion of topic keywords shown in the summary. \textbf{Speaker} $f_{Spk}$ is the number of tokens spoken by the selected speakers divided by the total number of tokens in the summary. \textbf{Length} $f_{Len}$ is the number of tokens in the summary. \textbf{Extractiveness} $f_{Ext}$ is the average of  ROUGE-2 precision and ROUGE-3 precision~\citep{lin-2004-rouge} of the generated summary against the input. For \textbf{Specificity}, inspired by previous studies~\cite{resnik1995using, amplayo2021aspect}, we find that verb, noun, numeral, and the total number of tokens show the most significant information about specificity. Thus, we define $f_{Spe}=(0.1 \times vb + 0.2 \times tok + 0.3 \times nn + 0.4 \times cd)/n_{s}$, where $vb$, $tok$, $nn$, $cd$, and $n_{s}$ represent the number of verbs, tokens, nouns, numeral tokens, and the number of sentences in the summary.

\subsection{Statistics of \ours}
\paragraph{Dataset Split and Source Data Distribution} 
Table~\ref{tab:statis} shows the statistics.
\ours~covers two domains (MAC-Doc for news and MAC-Dial for dialogue) with 8333 annotated summaries (5379 in MAC-Doc and 2954 in MAC-Dial), paired with 1353 source inputs (943 in MAC-Doc and 410 in MAC-Dial).
The averaged number of tokens in sources of MAC-Doc is shorter than that in the original QMSum dataset since we truncate the input into segments.
We split the source text randomly into training/valid/test sets with 80\%/10\%/10\%.

\begin{table}[t!]
\centering
\resizebox{\linewidth}{!}{
\begin{tabular}{@{}ccrrrrrr@{}}
\toprule
\multirow{2}{*}{Attribute}              &    \multirow{2}{*}{Value}                     & \multicolumn{3}{c}{MAC-Dial} & \multicolumn{3}{c}{MAC-Doc} \\ \cmidrule(rl){3-5} \cmidrule(rl){6-8} 
 &                   & \multicolumn{1}{c}{Train}    & \multicolumn{1}{c}{Dev}     & \multicolumn{1}{c}{Test}    & \multicolumn{1}{c}{Train}   & \multicolumn{1}{c}{Dev}     & \multicolumn{1}{c}{Test}    \\ \midrule
\multirow{3}{*}{Length}           & short                      & 38.04    & 39.52   & 43.84   & 31.97   & 33.37   & 34.30   \\
                                  & normal                      & 67.47    & 72.34   & 69.68   & 46.63   & 45.15   & 47.92   \\
                                  & long                      & 104.03   & 93.37   & 107.44  & 92.37   & 90.74   & 95.35   \\ \midrule
\multirow{3}{*}{Extractiveness}   & normal                      & 0.26     & 0.28    & 0.23    & 0.29    & 0.29    & 0.27    \\
                                  & high                      & 0.32     & 0.33    & 0.31    & 0.39    & 0.39    & 0.46    \\
                                  & full                      & 0.49     & 0.43    & 0.50    & 0.63    & 0.63    & 0.61    \\ \midrule
\multirow{2}{*}{Specificity}      & normal                      & 5.25     & 4.90    & 5.01    & 4.73    & 4.70    & 4.67    \\
                                  & high                      & 6.32     & 6.28    & 6.17    & 4.88    & 5.11    & 4.82    \\ \midrule
Topic                             & \multicolumn{1}{c}{--} & 0.83     & 0.81    & 0.79    & 0.95    & 0.98    & 0.95    \\ \midrule
Speaker                           & \multicolumn{1}{c}{--} & 0.74     & 0.71    & 0.71    & --      & --      & --      \\ \bottomrule
\end{tabular}}
\caption{Attribute metric functions $f_r$ of different control attribute values.
}
\label{tab:explore}
\end{table}

\paragraph{Distribution of Control Attribute Metrics}
With definitions in Section~\ref{sec:metrics}, Table~\ref{tab:explore} calculates automatic attribute metrics for all 5 control attributes.
As presented, the annotated summaries with different control attribute values can distinguish from each other by a large margin.
For example, samples with \textit{Len: long} have a much longer input, and samples with \textit{Ext: full} have a higher extractiveness metric. 
This verifies the high annotation quality of \ours~and also proves that our proposed attribute metrics are consistent with the control objective of each control attribute.

\begin{figure}[t!]
  \centering
  \includegraphics[width=3in]{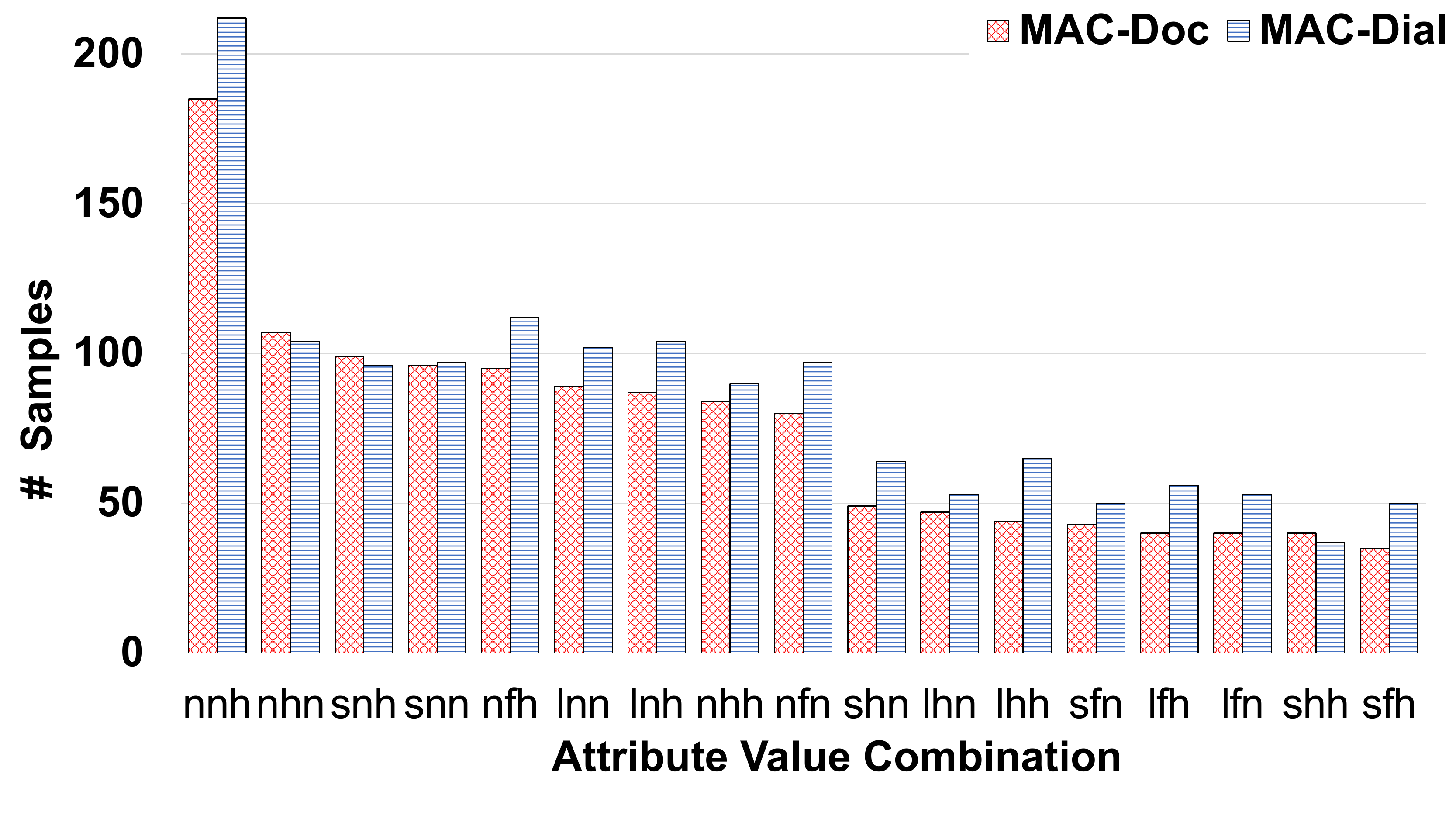}
  \caption{Distribution of mixed attributes. Each category is represented by the first character of its control attribute values, e.g., \textit{snh} represents \textit{Len: short}, \textit{Ext: normal}, and \textit{Spe: high}.}

  \label{fig:distri}
\end{figure}

\paragraph{Mixed-Attribute Distributions}
Figure~\ref{fig:distri} shows the ratio of different combinations of the control attributes. This illustrates diverse combinations of mixed-attributes summaries by controlling \textit{Len}, \textit{Ext}, and \textit{Spe} together in one sample.

\section{Methods}

\begin{figure*}[t!]
    \centering
    \includegraphics[width=\linewidth]{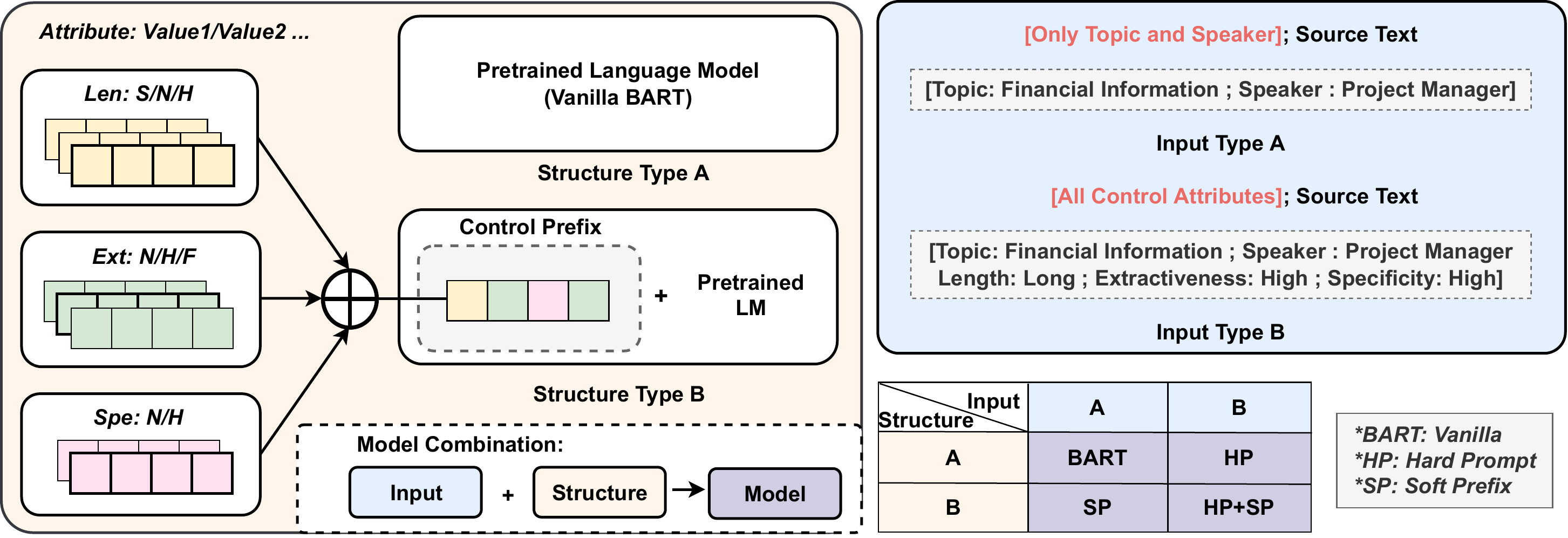}
    \caption{Comparison of different frameworks. For the HP model, the control attributes are prepended to the input to form a hard prompt. For the SP model, the selected prefix vectors are added together to form a control prefix. HP+SP contains both hard prompts and control prefixes.
    }
    \label{fig:prefix}
\end{figure*}

For setting baseline results on \ours, we propose three models following previous research on controllable text generation using prompt learning.
With the same input and different prompts, the large pretrained model is able to generate different results for different tasks, such as summarization and translation~\citep{he2020ctrlsum, fan2018controllable,raffel2020exploring}.
As shown in Figure~\ref{fig:prefix}, we leverage two types of prompt learning approaches to control the attributes of summaries, namely hard prompt (HP) and soft prefix tuning (SP). We also test the combination of them, namely HP+SP.

\paragraph{Hard Prompt (HP)} uses the description of control attributes as the hard prompt. Each attribute is formed as ``Attribute: Value'', where ``Attribute'' can be ``Topic, Speaker, Length, Extractiveness, Specificity'', and ``Value'' is the corresponding value (e.g., High or Normal) of the attribute. We concatenate 5 control attributes using ``;'' and prepend it to the input source.

\paragraph{Soft Prefix (SP)} follows~\citet{li-liang-2021-prefix}. We prepend external trainable parameters to both the encoder and decoder to control the summarization model. For controlling \textit{Len}, \textit{Ext}, and \textit{Spe}, we assign $m$ prefix embeddings for each attribute value where $m$ is a hyper-parameter meaning the length of prefix, i.e. prefix length. Readers can refer to~\citet{li-liang-2021-prefix} for implementation details.  For example, for \textit{Len: Long}, we assign $\bm{E}_{Len:long} = [\bm{e}^1_{Len:long}, \cdots, \bm{e}^m_{Len:long}]$ where $\bm{e}_i^j$ is a vector with dimension of word embedding.
And for controlling an input case with a set $\mathcal{V}$ of mixed requirements, we sum the embeddings of all control attribute values: $\bm{E} = [\sum_{v\in \mathcal{V}} \bm{e}^1_{v}, \cdots, \sum_{v\in \mathcal{V}} \bm{e}^m_{v}]$.
And for controlling \textit{Tpc} and  \textit{Spk}, we use the embeddings of input topics words $\bm{E}_{Tpc}$ and input speaker names $ \bm{E}_{Spk}$. This list of embedding vectors $\bm{E}$ is then prepended to each layer of the Transformer-based summarization model as external key/value vectors in its self-attention operations. $\bm{E}_{Tpc}$ and $\bm{E}_{Spk}$ are prepended only to the input layer.

\paragraph{Hard Prompt + Soft Prefix (HP+SP)} combines both approaches by prepending the hard prompt of five attributes in HP and using prefix tuning in SP.

\section{Experiments}
In this section, we present the implementation details, experimental results, and human evaluation of models on \ours~dataset.

\subsection{Implementation Details}
We use PyTorch and the Huggingface library~\citep{wolf2019huggingface} to implement our model. The experiments are conducted on 8 A100 GPUs.

We use BART~\citep{lewis-etal-2020-bart} as the backbone model. We also use a vanilla BART trained without control attribute input as a weak baseline (Appendix A). If not mentioned, we initialize the backbone using BART-large-cnn and then finetune it on the \ours~dataset. 
 We pick the $3e\text{-}5$ learning rate searching from $\{1e\text{-}5, 3e\text{-}5, 1e\text{-}4\}$ . Additionally, n-gram blocking is set to 3, and we use the AdamW optimizer with 500 warmup steps.
Dialogue inputs are flattened by separating turns with ``<$\backslash$s>'' which we find yields better results.

\begin{table*}[t!]
\normalsize
\centering
\resizebox{0.9\textwidth}{!}{

\begin{tabular}{@{}lrrrrrrrcrrr@{}}
\toprule
      & \multicolumn{2}{c}{Length}                       & \multicolumn{2}{c}{Extractiveness}               & \multicolumn{2}{c}{Specificity}                  & \multicolumn{1}{c}{Topic} & \multicolumn{1}{c}{Average} & \multicolumn{3}{c}{Quality}                                              \\ \cmidrule(lr){2-3} \cmidrule(lr){4-5}\cmidrule(lr){6-7} \cmidrule(lr){8-8} \cmidrule{9-9} \cmidrule(lr){10-12}
       & \multicolumn{1}{c}{CER$\downarrow$} & \multicolumn{1}{c}{CC$\updownarrow$} & \multicolumn{1}{c}{CER$\downarrow$} & \multicolumn{1}{c}{CC$\updownarrow$} & \multicolumn{1}{c}{CER$\downarrow$} & \multicolumn{1}{c}{CC$\updownarrow$} & \multicolumn{1}{c}{CER$\downarrow$} & \multicolumn{1}{c}{CER$\downarrow$} & \multicolumn{1}{c}{R1$\uparrow$} & \multicolumn{1}{c}{R2$\uparrow$} & \multicolumn{1}{c}{RL$\uparrow$} \\\midrule
      Gold   & 0.000  & 32.444  & 0.000  & 0.141   & 0.000  & 0.103  & 0.000  & 0.000    & 1.000  & 1.000   & 1.000 \\ \midrule
      BART  & 0.486  & 0.000  & 1.177  & 0.000  & 0.490  & 0.000  & 0.345 & 0.624  & 0.290    & 0.102    & 0.250  \\
      HP    & \textbf{0.340} & \textbf{31.421}  & \textbf{0.802}   & 0.239   & \textbf{0.353}   & 0.259  & \textbf{0.333}     & \textbf{0.457}       & \textbf{0.300}   & \textbf{0.104}  & \textbf{0.261} \\
      SP   & 0.475    & 4.671   & 1.111  & 0.055  & 0.466  & \textbf{0.105}   & 0.471    & 0.631   & 0.261    & 0.092 & 0.228 \\
      HP+SP & 0.373   & 25.226  & 1.136  & \textbf{0.133}  & 0.370   & 0.191     & 0.358    & 0.559    & 0.288   & 0.103 & 0.248  \\
 \bottomrule
\end{tabular}}
\caption{Results on MAC-Doc. The performance of the model is better when Control Error Rate (CER) is lower~$\downarrow$, ROUGE is higher~$\uparrow$, and Control Correlation (CC) is closer to the golden summary~$\updownarrow$.}
\label{tab:doc_results}
\end{table*}

\begin{table*}[t!]
\normalsize
\centering
\resizebox{\textwidth}{!}{
\begin{tabular}{@{}lrrrrrrrccrrr@{}}
\toprule
  & \multicolumn{2}{c}{Length}   & \multicolumn{2}{c}{Extractiveness}   & \multicolumn{2}{c}{Specificity}  & \multicolumn{1}{c}{Topic} & \multicolumn{1}{c}{Speaker} & \multicolumn{1}{c}{Average} & \multicolumn{3}{c}{Quality}  \\ \cmidrule(rl){2-3}\cmidrule(rl){4-5}\cmidrule(rl){6-7} \cmidrule(lr){8-8} \cmidrule(lr){9-9} \cmidrule(lr){10-10} \cmidrule(lr){11-13}
  & \multicolumn{1}{c}{CER$\downarrow$} & \multicolumn{1}{c}{CC$\updownarrow$} & \multicolumn{1}{c}{CER$\downarrow$} & \multicolumn{1}{c}{CC$\updownarrow$} & \multicolumn{1}{c}{CER$\downarrow$} & \multicolumn{1}{c}{CC$\updownarrow$} & \multicolumn{1}{c}{CER$\downarrow$}   & \multicolumn{1}{c}{CER$\downarrow$} & \multicolumn{1}{c}{CER$\downarrow$} & \multicolumn{1}{c}{R1$\uparrow$} & \multicolumn{1}{c}{R2$\uparrow$} & \multicolumn{1}{c}{RL$\uparrow$} \\\midrule
  Gold  & 0.000 & 42.045 & 0.000 & 0.088 & 0.000 & 1.610   & 0.000 & 0.000   & 0.000   & 1.000  & 1.000  & 1.000  \\\midrule
  BART  & 0.690 & 0.000 & 0.544 & 0.000  & 0.652 & 0.000  & 0.612 & 0.236   & 0.547   & \textbf{0.331}  & \textbf{0.113}  & \textbf{0.286}  \\
  HP & \textbf{0.577} & \textbf{12.629}  & 0.504 & \textbf{0.067} & \textbf{0.526} & \textbf{1.563} & \textbf{0.466}& \textbf{0.216}  & \textbf{0.458}  & 0.326 & 0.112 & 0.284  \\
  SP & 0.600 & -0.798 & \textbf{0.493} & 0.020 & 0.579  & 0.525  & 0.542 & 0.222   & 0.487   & 0.303  & 0.102  & 0.266  \\
  HP+SP  & 0.688 & -2.034  & 0.511 & 0.015 & 0.643  & 0.420  & 0.559 & 0.237   & 0.528   & 0.301  & 0.099  & 0.260  \\ \bottomrule
\end{tabular}}
\caption{Results on MAC-Dial. The performance of the model is better when Control Error Rate (CER) is lower~$\downarrow$, ROUGE is higher~$\uparrow$, and Control Correlation (CC) is closer to the golden summary~$\updownarrow$.}
\label{tab:dial_results}
\end{table*}
\subsection{Experiment Results}
As motioned in Section~\ref{sec:metrics}, we calculate Control Error Rate (CER) and Control Correlation (CC) metrics for evaluating control quality, and we also report ROUGE scores for evaluating summarization quality. For a model, its performance is better when the CER value is lower$\downarrow$, ROUGE is higher$\uparrow$, and its CC is closer to the golden summary$\updownarrow$. 

Table~\ref{tab:doc_results} shows the results of MAC-Doc. The HP model obtains the highest performance on both CER and CC across all 5 control attributes. Compared with the HP model, the SP model has similar control ability on \textit{Ext} and \textit{Spe}. However, it does not perform well on \textit{Len} and \textit{Tpc}. This could be the result of using the pretraining checkpoint that has learned some knowledge about the length-related hard prompt before training (Section~\ref{sec:pretrain}). 

Table~\ref{tab:dial_results} displays the results of MAC-Dial. Similar to the MAC-Doc dataset, the HP model obtains the highest scores on most of the metrics. However, the overall performance of length decreases because using the pretrained CNNDM checkpoint does not lead to performance gain in the dialogue domain (Section~\ref{sec:pretrain}). 

It is worth noting that the CER should not be compared across datasets, because its scale is different from different datasets.
For example, random uncontrolled BART in MAC-Doc obtains 1.177 CER for \textit{Ext} while it is 0.544 in MAC-Dial.

\subsection{Human Evaluation}
Although automatic metrics usually provide a speedy comparison, these metrics cannot easily evaluate the quality of the control, especially mixed-attribute control. Thus, we also conduct a human evaluation for the controlled summaries. 

\paragraph{Evaluation Method} We hire two evaluators with expertise in English and text summarization. We show them randomly-selected summaries generated by different systems with the source text and control attributes. The evaluators answer a yes/no question: ``For the given summary, does it follow the control requirement of this attribute?''. Specifically, we select golden summaries, summaries generated by HP model, and summaries generated by HP+SP model.
For each model, we pick 30 samples from MAC-Doc and MAC-Dial separately, resulting in 180 summaries in total. Furthermore, We compute Cohen's kappa~\citep{cohen1960coefficient} to measure the agreement between evaluators.

\begin{table}[t!]
\centering
\large
\resizebox{\linewidth}{!}{
\begin{tabular}{@{}lccccccccc@{}}
\toprule
      & \multicolumn{4}{c}{MAC-Doc} & \multicolumn{5}{c}{MAC-Dial}      \\  \cmidrule(lr){2-5} \cmidrule(lr){6-10}
      & \textit{Tpc}   & \textit{Ext}   & \textit{Spe}  & Kappa & \textit{Tpc}  & \textit{Spk}  & \textit{Ext}  & \textit{Spe}  & Kappa \\ \midrule
Gold  & 0.83  & 0.77  & 0.80  &  0.87   & 0.87   & 0.80  & 0.73  & 0.73  &  0.84     \\
HP    & 0.67  & 0.73  & 0.57  &  0.77   & 0.67   & 0.70  & 0.67  & 0.70  &  0.79       \\
HP+SP & 0.53  & 0.60  & 0.60  &  0.70   & 0.60   & 0.57  & 0.53  & 0.40  &  0.69      \\ \bottomrule
\end{tabular}}
\caption{Human evaluation results. We evaluate Speaker (\textit{Spk}), Extractiveness (\textit{Ext}), and Specificity (\textit{Spe}). Length does not require human annotation because it is measured by counting the number of tokens.}
\label{tab:human}
\end{table}

\paragraph{Evaluation Results} Table~\ref{tab:human} shows the human evaluation results. Each number (except for Kappa) is calculated by the count of yes answers divided by the total count of questions, indicating the control ability of the model. As shown, the HP model performs better than HP+SP on most of the attributes. This result confirms the consistency of our proposed CER and CC with human evaluation.

Besides, golden summaries always rank first, and the kappa score of the two evaluators is over 0.8. These two results also verify the high annotation quality of~\ours, because human evaluators agreed that the golden summaries followed the control requirements most of the time.
\section{Analysis and Discussion}
For a deeper understanding of the task of mixed-attribute controllable summarization on~\ours, we conduct analysis including attribute difficulty, attribute dependency, model pretraining, and present several example outputs for case studies.

\subsection{Difficulty of Controlling Attribute Values}

\begin{figure}[t!]
  \centering
  \includegraphics[width=\linewidth]{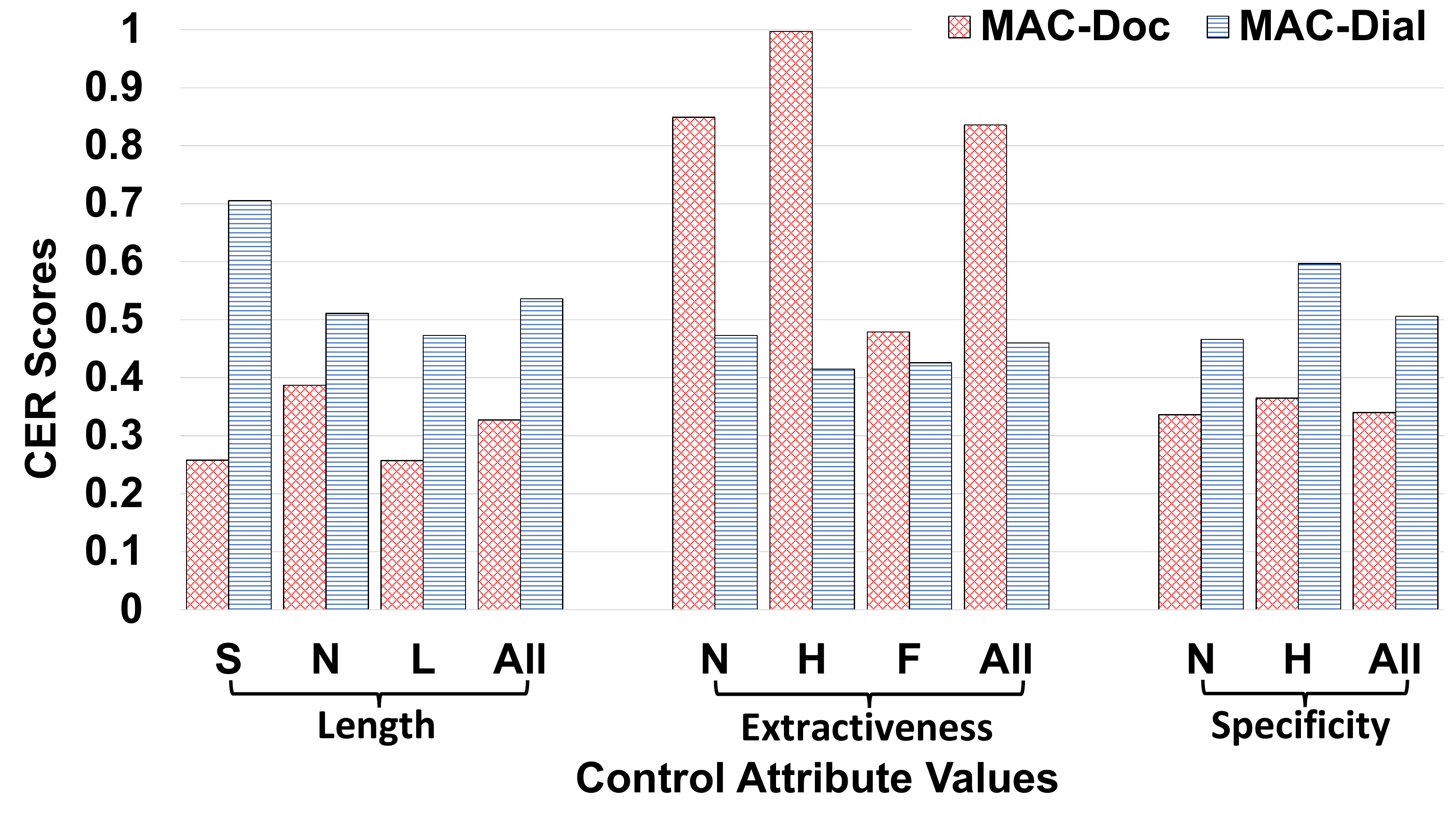}
  \caption{Difficulty of attribute values. The x-axis shows the control attribute and its value. For instance, S in length is the CER of all the \textit{Len: short} samples.}
  \label{fig:difficulty}
\end{figure}

Models have different difficulties in controlling certain attribute values, as some attribute values can be easier or harder to be controlled. We analyze this by comparing CER for different attribute values of the HP model's outputs.
As shown in Figure~\ref{fig:difficulty}, for MAC-Doc, the system obtains a higher CER on \textit{Len: normal} samples compared with the other two values of \textit{Len}, showing that \textit{normal} is more difficult to control, and the hardest values in controlling \textit{Ext} and \textit{Spe} are both \textit{high}. For MAC-Dial, the hardest values in controlling \textit{Len}, \textit{Ext}, and \textit{Spe} are \textit{short}, \textit{normal}, and \textit{high} respectively.

\subsection{Dependency of Attributes}
\begin{figure}
  \centering
\subfigure[MAC-Doc attribute dependency]{\includegraphics[width=0.87\linewidth]{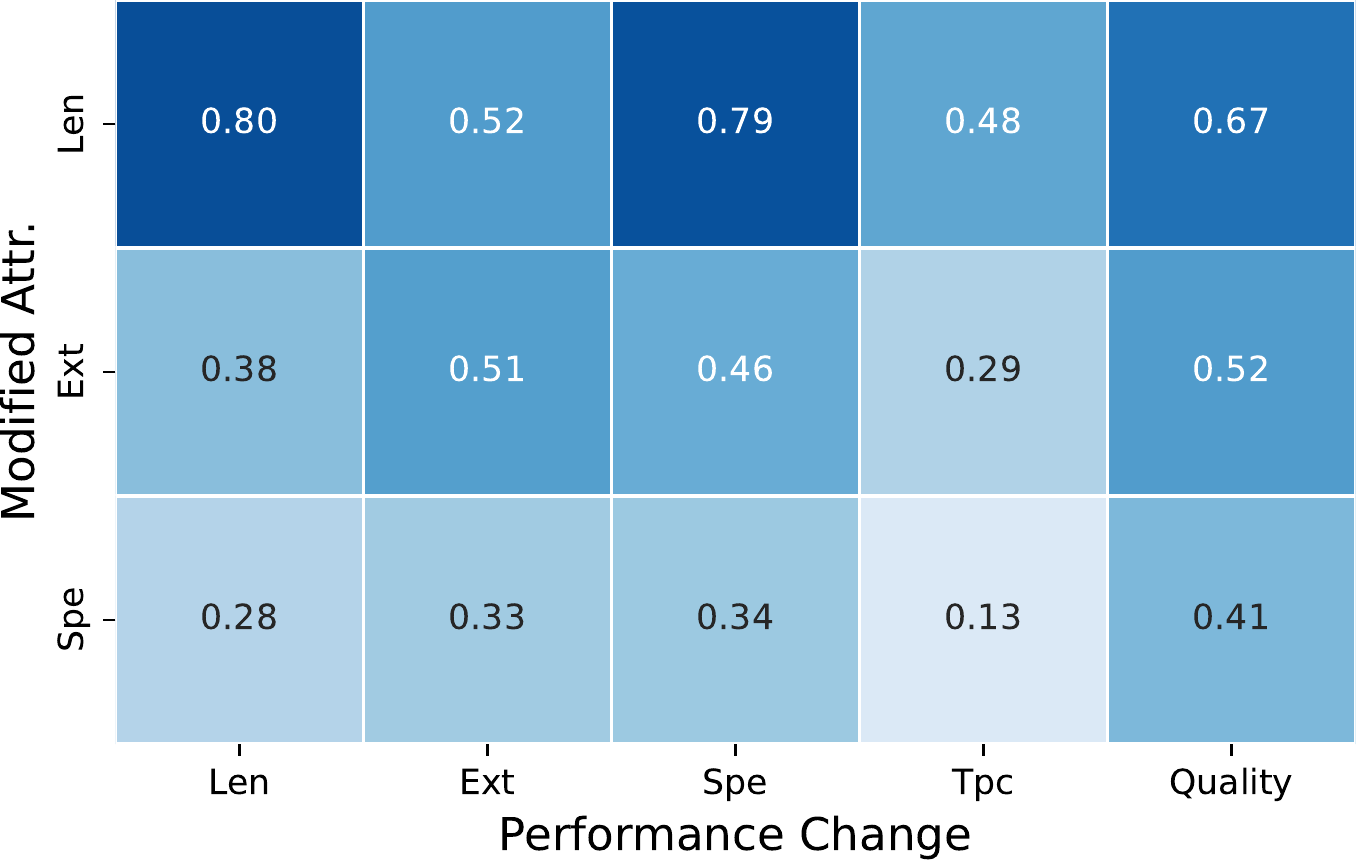}}\hspace{5mm}
\subfigure[MAC-Dial attribute dependency]{\includegraphics[width=\linewidth]{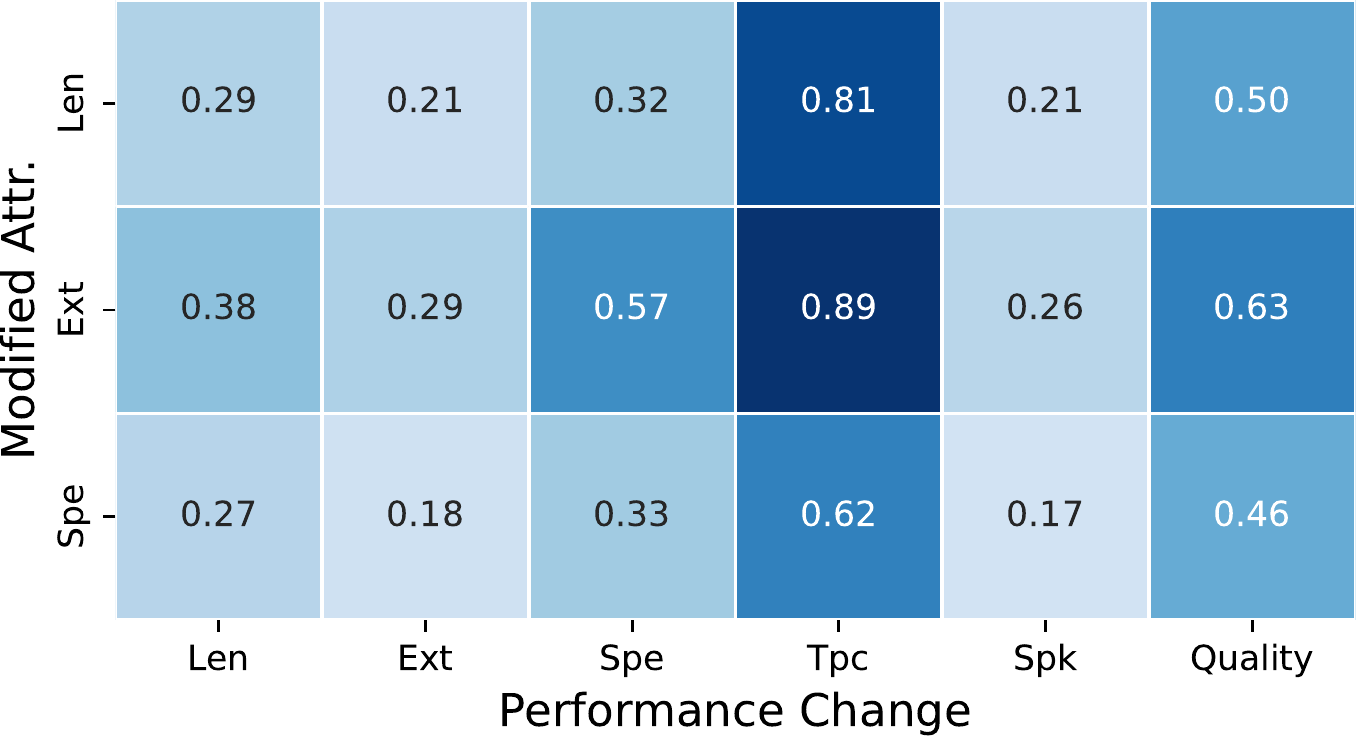}}
  
  \caption{Dependency of attributes. Each row shows the attributes that are modified while each column shows the change in the corresponding attribute.}
  \label{fig:dependency}
\end{figure}
In mixed-attribute controllable summarization, we notice interesting dependencies among attributes, as changing one attribute influences the other one.
To analyze this, we randomly select 200 samples from the test set for each attribute, and randomly change this attribute to another value to form a new sample (e.g., from \textit{Len: long} to \textit{Len: short}). Then, the same HP model, without further training, is used to generate summaries on these new samples. We evaluate the performance difference between the newly predicted summaries $\hat{y}'$ and the originally predicted summaries $\hat{y}$ via $\text{CER}(\hat{y}',\hat{y})$.

Figure~\ref{fig:dependency} shows the performance change. As can be seen, for MAC-Doc, \textit{Len} has the highest dependency toward other attributes, while \textit{Spe} has the lowest. For MAC-Dial, \textit{Ext} has the highest dependency, while \textit{Spe} has the lowest. We believe this is because the model in MAC-Doc has a strong control ability towards \textit{Len}. Thus, the value change of \textit{Len} will influence more on other attributes.

\begin{table}[t!]
\large
\resizebox{\linewidth}{!}{
\begin{tabular}{@{}lcccccc@{}}
\toprule
         & Len   & Ext   & Spe   & Tpc   & Spk   & Quality \\ \midrule
MAC-Doc      &  0.315    & 0.870      & 0.327      & 0.254   & --  &  0.776   \\
\; -CNN      &  0.361    &  1.033     &  0.392     & 0.346   & --  &  0.777       \\
\midrule
MAC-Dial & 0.454 & 0.392 & 0.420 & 0.373 & 0.211 & 0.719   \\
\; -CNN      & 0.469 & 0.422 & 0.430 & 0.476 & 0.201 & 0.722   \\
\bottomrule
\end{tabular}}
\caption{Ablation on \ours~on pretraining on CNNDM. MAC-Doc, MAC-Dial denote the HP model initialized with BART-large-cnn, while -CNN uses BART-large checkpoint. Numbers for five control attributes are CER and for Quality are the average of ROUGE-1/2/L.}
\label{tab:ablation}
\end{table}
\subsection{Effect of Pretraining}
\label{sec:pretrain}
We investigate the effect of pretraining on the control ability of summarization models.
For two HP models initialized by BART-large and BART-large-cnn separately, we compare their results after finetuning them on both MAC-Doc and MAC-Dial.

As shown in Table~\ref{tab:ablation}, for MAC-Doc, the BART-large-cnn initialized model is able to control the length substantially better than the vanilla BART-large initialized model.
On the contrary, for MAC-Dial, the advantage of the BART-large-cnn checkpoint is negligible. Using BART-large-cnn or not only slightly influences the control ability of all attributes in MAC-Dial. We believe the reason for this is that the CNNDM pretraining provides certain useful information for the model to learn the ability to control attributes on news articles.

\begin{table*}[!t]
    \centering
    \small
    \begin{tabular}{lp{13cm}}
    \Xhline{5\arrayrulewidth}
    \multicolumn{2}{c}{Case 1: Topic Defocus (MAC-Doc)}\\
    \midrule
        Attributes  &  Length : normal ; Extractiveness : full ; Specificity : normal ; \textbf{Topic : education}  \\
        \midrule
        Gold & They quickly reopened the University of Mosul, under a radically altered curriculum. Some subjects would be banned -- democracy and political thought, hotel management, tourism and archaeology. ISIS allows girls to go to school, in a segregated environment.\\
        \midrule
        HP & The Taliban, forbids all girls' education. But ISIS allows girls to go to school, albeit in a segregated environment. \\
    \Xhline{5\arrayrulewidth}
    \multicolumn{2}{c}{Case 2: Length against Specificity (MAC-Doc)}\\
    \midrule
        Attributes   &  \textbf{Length : short} ; Extractiveness : normal ; \textbf{Specificity : high} ; Topic: campaign  \\
        \midrule
        Gold & All variations of women feature in Sunday's campaign video release by Hillary Clinton. \\
        \midrule
        HP & Clinton's new campaign website is focused on women and the progress of women in politics.\\
        \midrule
        HP+SP & During her campaign, Hillary Clinton has focused on women's issues. Her new campaign website is filled with women's pictures. \\
    \Xhline{5\arrayrulewidth}
    \multicolumn{2}{c}{Case 3: Extractiveness against Readability (MAC-Dial)}\\
    \midrule
        Attributes   &  Length : normal ; \textbf{Extractiveness : full} ; Specificity : normal ; Topic : parallel marketing , angle ; Speaker : Marketing  \\
        \midrule
        Gold & Marketing; either market it as the point of view; we could have parallel marketing s schemes; one where you've got one where it appeals to people that want to have the new device that looks cool, is fashionable; So um, I dunno we'll have to decide which which angle we're gonna go to or both; Either market it together by getting control in a set colour or like you buy it with several; as a separate thing. \\
        \midrule
        HP+SP & Marketing, could have parallel marketing, schemes, one where it appeals to people that want to have the new device that looks cool; one that rather, than a kind of a need relationship with the device; people might not like, having a device, just looks nice; also a device, practically sound; decide which angle, gonna go to or both. \\
    \Xhline{5\arrayrulewidth}
    \end{tabular}
    \caption{Three case studies on \ours.}
    \label{tab:case_study}
\end{table*}

\subsection{Case Study}
We show three case studies in Table~\ref{tab:case_study}, discussing three typical phenomena in mixed-attribute controllable summarization, namely Topic Defocus, Length against Specificity, and Extractiveness against Readability.

\paragraph{Topic Defocus} In Table~\ref{tab:case_study} Case 1, \ours~asks for a summary focusing on the topic of ``\textit{education}''.
Although the human-annotated summary does not contain the topic word, its contents are still highly related to ``\textit{education}''.
This shows that human annotators have the flexibility of conducting high-level summarization of the topic.
In contrast, although the model-generated summary contains the topic word, its content is poorly structured.
This shows the challenge of topic defocus, a phenomenon where models rely too much on explicitly containing the topic words when generating topic-controlled summaries.

\paragraph{Length against Specificity} Another challenge is the contradiction between long length and low specificity. Long summaries contain more tokens and inevitably invite more specific information. On the contrary, short summaries only describe core events using a few words and are naturally biased towards low specificity. As shown in Table~\ref{tab:case_study} Case 2, when \textit{Len} is \textit{short} and \textit{Spe} is \textit{high}, both HP and HP+SP generated summaries are longer compared with the human-annotated summary.

\paragraph{Extractiveness against Readability} As shown in Table~\ref{tab:case_study} Case 3, when \textit{Ext} is \textit{full}, the model-generated summaries are choppy and unnatural, in particular for dialogues.
When humans are asked to annotate fully extractive summaries, they may have to write unnatural sentences, and this phenomenon is amplified by a trained summarization system. 
As shown in the table, the HP+SP generated summary is not grammatical and consists of short phrases instead of complete sentences.
This can be explained by the fact that the complicated dialogue discourse structures and frequent interactions between different interlocutors make salient information sparse.

\section{Conclusion}
We propose \ours, a high-quality human-annotated benchmark for mixed-attribute controllable summarization. It contains 5 types of control attributes, including Topic, Speaker, Length, Extractiveness, and Specificity. To the best of our knowledge, \ours~is the first dataset with mixed attributes as well as human annotations. We explore the hard prompt and soft prefix models and evaluate them on \ours. Results and analysis demonstrate that hard prompt models yield the best performance and also show this is a challenging task as a large gap between machine learning models and human still exists. 

Future work can design more effective models 
 for the mixed-attribute controllable summarization task, or explore mixed-attribute control on other generation tasks. 

\section*{Acknowledgments}
We thank Ming Zhong, Tao Yu, Haoran Zhang, Sarkar Das, Nan Zhang for their valuable feedback and suggestions. We also would like to thank the reviewers and action editors for their helpful comments and reviews.

\bibliography{tacl2021}

\begin{thebibliography}{53}
\expandafter\ifx\csname natexlab\endcsname\relax\def\natexlab#1{#1}\fi

\bibitem[{Amplayo et~al.(2021)Amplayo, Angelidis, and
  Lapata}]{amplayo2021aspect}
Reinald~Kim Amplayo, Stefanos Angelidis, and Mirella Lapata. 2021.
\newblock \href {https://doi.org/10.18653/v1/2021.emnlp-main.528}
  {Aspect-controllable opinion summarization}.
\newblock In \emph{Proceedings of the 2021 Conference on Empirical Methods in
  Natural Language Processing}, pages 6578--6593, Online and Punta Cana,
  Dominican Republic. Association for Computational Linguistics.

\bibitem[{B{\"o}hm et~al.(2019)B{\"o}hm, Gao, Meyer, Shapira, Dagan, and
  Gurevych}]{bohm2019better}
Florian B{\"o}hm, Yang Gao, Christian~M. Meyer, Ori Shapira, Ido Dagan, and
  Iryna Gurevych. 2019.
\newblock \href {https://doi.org/10.18653/v1/D19-1307} {Better rewards yield
  better summaries: Learning to summarise without references}.
\newblock In \emph{Proceedings of the 2019 Conference on Empirical Methods in
  Natural Language Processing and the 9th International Joint Conference on
  Natural Language Processing (EMNLP-IJCNLP)}, pages 3110--3120, Hong Kong,
  China. Association for Computational Linguistics.

\bibitem[{Bornstein et~al.(1999)Bornstein, Cutting, Hatton, and
  Rose}]{bornstein1999interactive}
Jeremy~J Bornstein, Douglass~R Cutting, John~D Hatton, and Daniel~E Rose. 1999.
\newblock Interactive document summarization.
\newblock US Patent 5,867,164.

\bibitem[{Boudin(2016)}]{boudin:2016:COLINGDEMO}
Florian Boudin. 2016.
\newblock \href {https://aclanthology.org/C16-2015} {pke: an open source
  python-based keyphrase extraction toolkit}.
\newblock In \emph{Proceedings of {COLING} 2016, the 26th International
  Conference on Computational Linguistics: System Demonstrations}, pages
  69--73, Osaka, Japan. The COLING 2016 Organizing Committee.

\bibitem[{Brown et~al.(2020)Brown, Mann, Ryder, Subbiah, Kaplan, Dhariwal,
  Neelakantan, Shyam, Sastry, Askell, Agarwal, Herbert{-}Voss, Krueger,
  Henighan, Child, Ramesh, Ziegler, Wu, Winter, Hesse, Chen, Sigler, Litwin,
  Gray, Chess, Clark, Berner, McCandlish, Radford, Sutskever, and
  Amodei}]{DBLP:conf/nips/BrownMRSKDNSSAA20}
Tom~B. Brown, Benjamin Mann, Nick Ryder, Melanie Subbiah, Jared Kaplan,
  Prafulla Dhariwal, Arvind Neelakantan, Pranav Shyam, Girish Sastry, Amanda
  Askell, Sandhini Agarwal, Ariel Herbert{-}Voss, Gretchen Krueger, Tom
  Henighan, Rewon Child, Aditya Ramesh, Daniel~M. Ziegler, Jeffrey Wu, Clemens
  Winter, Christopher Hesse, Mark Chen, Eric Sigler, Mateusz Litwin, Scott
  Gray, Benjamin Chess, Jack Clark, Christopher Berner, Sam McCandlish, Alec
  Radford, Ilya Sutskever, and Dario Amodei. 2020.
\newblock \href
  {https://proceedings.neurips.cc/paper/2020/hash/1457c0d6bfcb4967418bfb8ac142f64a-Abstract.html}
  {Language models are few-shot learners}.
\newblock In \emph{Advances in Neural Information Processing Systems 33: Annual
  Conference on Neural Information Processing Systems 2020, NeurIPS 2020,
  December 6-12, 2020, virtual}.

\bibitem[{Cao and Wang(2021)}]{cao2021inference}
Shuyang Cao and Lu~Wang. 2021.
\newblock \href {https://doi.org/10.18653/v1/2021.naacl-main.476} {Inference
  time style control for summarization}.
\newblock In \emph{Proceedings of the 2021 Conference of the North American
  Chapter of the Association for Computational Linguistics: Human Language
  Technologies}, pages 5942--5953, Online. Association for Computational
  Linguistics.

\bibitem[{Chan et~al.(2021)Chan, Wang, and King}]{chan2021controllable}
Hou~Pong Chan, Lu~Wang, and Irwin King. 2021.
\newblock \href {https://doi.org/10.1162/tacl_a_00423} {Controllable
  summarization with constrained {M}arkov decision process}.
\newblock \emph{Transactions of the Association for Computational Linguistics},
  9:1213--1232.

\bibitem[{Chen et~al.(2022)Chen, Liu, Dong, Wang, Zhu, Zeng, and
  Zhang}]{DBLP:journals/corr/abs-2202-04824}
Yulong Chen, Yang Liu, Li~Dong, Shuohang Wang, Chenguang Zhu, Michael Zeng, and
  Yue Zhang. 2022.
\newblock \href {https://aclanthology.org/2022.findings-emnlp.448}
  {{A}da{P}rompt: Adaptive model training for prompt-based {NLP}}.
\newblock In \emph{Findings of the Association for Computational Linguistics:
  EMNLP 2022}, pages 6057--6068, Abu Dhabi, United Arab Emirates. Association
  for Computational Linguistics.

\bibitem[{Cheng and Lapata(2016)}]{cheng2016neural}
Jianpeng Cheng and Mirella Lapata. 2016.
\newblock \href {https://doi.org/10.18653/v1/P16-1046} {Neural summarization by
  extracting sentences and words}.
\newblock In \emph{Proceedings of the 54th Annual Meeting of the Association
  for Computational Linguistics (Volume 1: Long Papers)}, pages 484--494,
  Berlin, Germany. Association for Computational Linguistics.

\bibitem[{Cohen(1960)}]{cohen1960coefficient}
Jacob Cohen. 1960.
\newblock A coefficient of agreement for nominal scales.
\newblock \emph{Educational and psychological measurement}, 20(1):37--46.

\bibitem[{Dang(2005)}]{dang2005overview}
Hoa~Trang Dang. 2005.
\newblock Overview of duc 2005.
\newblock In \emph{Proceedings of the document understanding conference},
  volume 2005, pages 1--12.

\bibitem[{Daum{\'e}~III and Marcu(2006)}]{daume2009bayesian}
Hal Daum{\'e}~III and Daniel Marcu. 2006.
\newblock \href {https://doi.org/10.3115/1220175.1220214} {{B}ayesian
  query-focused summarization}.
\newblock In \emph{Proceedings of the 21st International Conference on
  Computational Linguistics and 44th Annual Meeting of the Association for
  Computational Linguistics}, pages 305--312, Sydney, Australia. Association
  for Computational Linguistics.

\bibitem[{Dou et~al.(2021)Dou, Liu, Hayashi, Jiang, and
  Neubig}]{dou-etal-2021-gsum}
Zi-Yi Dou, Pengfei Liu, Hiroaki Hayashi, Zhengbao Jiang, and Graham Neubig.
  2021.
\newblock \href {https://doi.org/10.18653/v1/2021.naacl-main.384} {{GS}um: A
  general framework for guided neural abstractive summarization}.
\newblock In \emph{Proceedings of the 2021 Conference of the North American
  Chapter of the Association for Computational Linguistics: Human Language
  Technologies}, pages 4830--4842, Online. Association for Computational
  Linguistics.

\bibitem[{Erkan and Radev(2004)}]{erkan2004lexrank}
G{\"u}nes Erkan and Dragomir~R Radev. 2004.
\newblock Lexrank: Graph-based lexical centrality as salience in text
  summarization.
\newblock \emph{Journal of artificial intelligence research}, 22:457--479.

\bibitem[{Fan et~al.(2018)Fan, Grangier, and Auli}]{fan2018controllable}
Angela Fan, David Grangier, and Michael Auli. 2018.
\newblock \href {https://doi.org/10.18653/v1/W18-2706} {Controllable
  abstractive summarization}.
\newblock In \emph{Proceedings of the 2nd Workshop on Neural Machine
  Translation and Generation}, pages 45--54, Melbourne, Australia. Association
  for Computational Linguistics.

\bibitem[{Fisher and Roark(2006)}]{fisher2006query}
Seeger Fisher and Brian Roark. 2006.
\newblock Query-focused summarization by supervised sentence ranking and skewed
  word distributions.
\newblock In \emph{Proceedings of the Document Understanding Conference,
  DUC-2006, New York, USA}. Citeseer.

\bibitem[{Goyal et~al.(2022{\natexlab{a}})Goyal, Li, and
  Durrett}]{goyal2022news}
Tanya Goyal, Junyi~Jessy Li, and Greg Durrett. 2022{\natexlab{a}}.
\newblock \href {https://arxiv.org/abs/2209.12356} {News summarization and
  evaluation in the era of gpt-3}.
\newblock \emph{ArXiv preprint}, abs/2209.12356.

\bibitem[{Goyal et~al.(2022{\natexlab{b}})Goyal, Rajani, Liu, and
  Kryscinski}]{goyal2021hydrasum}
Tanya Goyal, Nazneen Rajani, Wenhao Liu, and Wojciech Kryscinski.
  2022{\natexlab{b}}.
\newblock \href {https://aclanthology.org/2022.emnlp-main.30} {{H}ydra{S}um:
  Disentangling style features in text summarization with multi-decoder
  models}.
\newblock In \emph{Proceedings of the 2022 Conference on Empirical Methods in
  Natural Language Processing}, pages 464--479, Abu Dhabi, United Arab
  Emirates. Association for Computational Linguistics.

\bibitem[{He et~al.(2020)He, Kry{\'s}ci{\'n}ski, McCann, Rajani, and
  Xiong}]{he2020ctrlsum}
Junxian He, Wojciech Kry{\'s}ci{\'n}ski, Bryan McCann, Nazneen Rajani, and
  Caiming Xiong. 2020.
\newblock \href {https://arxiv.org/abs/2012.04281} {Ctrlsum: Towards generic
  controllable text summarization}.
\newblock \emph{ArXiv preprint}, abs/2012.04281.

\bibitem[{He et~al.(2022)He, Peng, Lu, Wang, Mei, Liu, Xu, Awadalla, Shi, Zhu
  et~al.}]{he2022z}
Pengcheng He, Baolin Peng, Liyang Lu, Song Wang, Jie Mei, Yang Liu, Ruochen Xu,
  Hany~Hassan Awadalla, Yu~Shi, Chenguang Zhu, et~al. 2022.
\newblock \href {https://arxiv.org/abs/2208.09770} {Z-code++: A pre-trained
  language model optimized for abstractive summarization}.
\newblock \emph{ArXiv preprint}, abs/2208.09770.

\bibitem[{Hermann et~al.(2015)Hermann, Kocisk{\'{y}}, Grefenstette, Espeholt,
  Kay, Suleyman, and Blunsom}]{NIPS2015_afdec700}
Karl~Moritz Hermann, Tom{\'{a}}s Kocisk{\'{y}}, Edward Grefenstette, Lasse
  Espeholt, Will Kay, Mustafa Suleyman, and Phil Blunsom. 2015.
\newblock \href
  {https://proceedings.neurips.cc/paper/2015/hash/afdec7005cc9f14302cd0474fd0f3c96-Abstract.html}
  {Teaching machines to read and comprehend}.
\newblock In \emph{Advances in Neural Information Processing Systems 28: Annual
  Conference on Neural Information Processing Systems 2015, December 7-12,
  2015, Montreal, Quebec, Canada}, pages 1693--1701.

\bibitem[{Hofmann-Coyle et~al.(2022)Hofmann-Coyle, Kulkarni, Xie, Maddela, and
  Preotiuc-Pietro}]{hofmann2022extractive}
Ella Hofmann-Coyle, Mayank Kulkarni, Lingjue Xie, Mounica Maddela, and Daniel
  Preotiuc-Pietro. 2022.
\newblock Extractive entity-centric summarization as sentence selection using
  bi-encoders.
\newblock In \emph{Proceedings of the 2nd Conference of the Asia-Pacific
  Chapter of the Association for Computational Linguistics and the 12th
  International Joint Conference on Natural Language Processing (Volume 2:
  Short Papers)}.

\bibitem[{Lester et~al.(2021)Lester, Al-Rfou, and
  Constant}]{lester-etal-2021-power}
Brian Lester, Rami Al-Rfou, and Noah Constant. 2021.
\newblock \href {https://doi.org/10.18653/v1/2021.emnlp-main.243} {The power of
  scale for parameter-efficient prompt tuning}.
\newblock In \emph{Proceedings of the 2021 Conference on Empirical Methods in
  Natural Language Processing}, pages 3045--3059, Online and Punta Cana,
  Dominican Republic. Association for Computational Linguistics.

\bibitem[{Leuski et~al.(2003)Leuski, Lin, and Hovy}]{leuski2003ineats}
Anton Leuski, Chin-Yew Lin, and Eduard Hovy. 2003.
\newblock \href {https://doi.org/10.3115/1075178.1075197} {i{N}e{ATS}:
  Interactive multi-document summarization}.
\newblock In \emph{The Companion Volume to the Proceedings of 41st Annual
  Meeting of the Association for Computational Linguistics}, pages 125--128,
  Sapporo, Japan. Association for Computational Linguistics.

\bibitem[{Lewis et~al.(2020)Lewis, Liu, Goyal, Ghazvininejad, Mohamed, Levy,
  Stoyanov, and Zettlemoyer}]{lewis-etal-2020-bart}
Mike Lewis, Yinhan Liu, Naman Goyal, Marjan Ghazvininejad, Abdelrahman Mohamed,
  Omer Levy, Veselin Stoyanov, and Luke Zettlemoyer. 2020.
\newblock \href {https://doi.org/10.18653/v1/2020.acl-main.703} {{BART}:
  Denoising sequence-to-sequence pre-training for natural language generation,
  translation, and comprehension}.
\newblock In \emph{Proceedings of the 58th Annual Meeting of the Association
  for Computational Linguistics}, pages 7871--7880, Online. Association for
  Computational Linguistics.

\bibitem[{Li and Liang(2021)}]{li-liang-2021-prefix}
Xiang~Lisa Li and Percy Liang. 2021.
\newblock \href {https://doi.org/10.18653/v1/2021.acl-long.353} {Prefix-tuning:
  Optimizing continuous prompts for generation}.
\newblock In \emph{Proceedings of the 59th Annual Meeting of the Association
  for Computational Linguistics and the 11th International Joint Conference on
  Natural Language Processing (Volume 1: Long Papers)}, pages 4582--4597,
  Online. Association for Computational Linguistics.

\bibitem[{Lin(2004)}]{lin-2004-rouge}
Chin-Yew Lin. 2004.
\newblock \href {https://aclanthology.org/W04-1013} {{ROUGE}: A package for
  automatic evaluation of summaries}.
\newblock In \emph{Text Summarization Branches Out}, pages 74--81, Barcelona,
  Spain. Association for Computational Linguistics.

\bibitem[{Liu et~al.(2021)Liu, Zheng, Du, Ding, Qian, Yang, and
  Tang}]{liu2021gpt}
Xiao Liu, Yanan Zheng, Zhengxiao Du, Ming Ding, Yujie Qian, Zhilin Yang, and
  Jie Tang. 2021.
\newblock \href {https://arxiv.org/abs/2103.10385} {Gpt understands, too}.
\newblock \emph{ArXiv preprint}, abs/2103.10385.

\bibitem[{Liu et~al.(2022)Liu, Jia, and Zhu}]{liu2022length}
Yizhu Liu, Qi~Jia, and Kenny Zhu. 2022.
\newblock \href {https://doi.org/10.18653/v1/2022.acl-long.474} {Length control
  in abstractive summarization by pretraining information selection}.
\newblock In \emph{Proceedings of the 60th Annual Meeting of the Association
  for Computational Linguistics (Volume 1: Long Papers)}, pages 6885--6895,
  Dublin, Ireland. Association for Computational Linguistics.

\bibitem[{Liu et~al.(2018)Liu, Luo, and Zhu}]{liu2018controlling}
Yizhu Liu, Zhiyi Luo, and Kenny Zhu. 2018.
\newblock \href {https://doi.org/10.18653/v1/D18-1444} {Controlling length in
  abstractive summarization using a convolutional neural network}.
\newblock In \emph{Proceedings of the 2018 Conference on Empirical Methods in
  Natural Language Processing}, pages 4110--4119, Brussels, Belgium.
  Association for Computational Linguistics.

\bibitem[{Louis and Nenkova(2011)}]{louis2011text}
Annie Louis and Ani Nenkova. 2011.
\newblock \href {https://aclanthology.org/W11-1605} {Text specificity and
  impact on quality of news summaries}.
\newblock In \emph{Proceedings of the Workshop on Monolingual Text-To-Text
  Generation}, pages 34--42, Portland, Oregon. Association for Computational
  Linguistics.

\bibitem[{Maddela et~al.(2022)Maddela, Kulkarni, and
  Preotiuc-Pietro}]{maddela2022entsum}
Mounica Maddela, Mayank Kulkarni, and Daniel Preotiuc-Pietro. 2022.
\newblock \href {https://doi.org/10.18653/v1/2022.acl-long.237} {{E}nt{SUM}: A
  data set for entity-centric extractive summarization}.
\newblock In \emph{Proceedings of the 60th Annual Meeting of the Association
  for Computational Linguistics (Volume 1: Long Papers)}, pages 3355--3366,
  Dublin, Ireland. Association for Computational Linguistics.

\bibitem[{Makino et~al.(2019)Makino, Iwakura, Takamura, and
  Okumura}]{makino2019global}
Takuya Makino, Tomoya Iwakura, Hiroya Takamura, and Manabu Okumura. 2019.
\newblock \href {https://doi.org/10.18653/v1/P19-1099} {Global optimization
  under length constraint for neural text summarization}.
\newblock In \emph{Proceedings of the 57th Annual Meeting of the Association
  for Computational Linguistics}, pages 1039--1048, Florence, Italy.
  Association for Computational Linguistics.

\bibitem[{McKeown and Radev(1995)}]{mckeown1995generating}
Kathleen McKeown and Dragomir~R Radev. 1995.
\newblock Generating summaries of multiple news articles.
\newblock In \emph{Proceedings of the 18th annual international ACM SIGIR
  conference on Research and development in information retrieval}, pages
  74--82.

\bibitem[{Min et~al.(2022)Min, Lewis, Hajishirzi, and
  Zettlemoyer}]{DBLP:conf/acl/MinLHZ22}
Sewon Min, Mike Lewis, Hannaneh Hajishirzi, and Luke Zettlemoyer. 2022.
\newblock \href {https://doi.org/10.18653/v1/2022.acl-long.365} {Noisy channel
  language model prompting for few-shot text classification}.
\newblock In \emph{Proceedings of the 60th Annual Meeting of the Association
  for Computational Linguistics (Volume 1: Long Papers)}, pages 5316--5330,
  Dublin, Ireland. Association for Computational Linguistics.

\bibitem[{Narayan et~al.(2021)Narayan, Zhao, Maynez, Sim{\~o}es, Nikolaev, and
  McDonald}]{narayan2021planning}
Shashi Narayan, Yao Zhao, Joshua Maynez, Gon{\c{c}}alo Sim{\~o}es, Vitaly
  Nikolaev, and Ryan McDonald. 2021.
\newblock \href {https://doi.org/10.1162/tacl_a_00438} {Planning with learned
  entity prompts for abstractive summarization}.
\newblock \emph{Transactions of the Association for Computational Linguistics},
  9:1475--1492.

\bibitem[{Paulus et~al.(2018)Paulus, Xiong, and Socher}]{paulus2018deep}
Romain Paulus, Caiming Xiong, and Richard Socher. 2018.
\newblock \href {https://openreview.net/forum?id=HkAClQgA-} {A deep reinforced
  model for abstractive summarization}.
\newblock In \emph{6th International Conference on Learning Representations,
  {ICLR} 2018, Vancouver, BC, Canada, April 30 - May 3, 2018, Conference Track
  Proceedings}. OpenReview.net.

\bibitem[{Qin and Eisner(2021)}]{DBLP:conf/naacl/QinE21}
Guanghui Qin and Jason Eisner. 2021.
\newblock \href {https://doi.org/10.18653/v1/2021.naacl-main.410} {Learning how
  to ask: Querying {LM}s with mixtures of soft prompts}.
\newblock In \emph{Proceedings of the 2021 Conference of the North American
  Chapter of the Association for Computational Linguistics: Human Language
  Technologies}, pages 5203--5212, Online. Association for Computational
  Linguistics.

\bibitem[{Raffel et~al.(2020)Raffel, Shazeer, Roberts, Lee, Narang, Matena,
  Zhou, Li, Liu et~al.}]{raffel2020exploring}
Colin Raffel, Noam Shazeer, Adam Roberts, Katherine Lee, Sharan Narang, Michael
  Matena, Yanqi Zhou, Wei Li, Peter~J Liu, et~al. 2020.
\newblock Exploring the limits of transfer learning with a unified text-to-text
  transformer.
\newblock \emph{J. Mach. Learn. Res.}, 21(140):1--67.

\bibitem[{Resnik(1995)}]{resnik1995using}
Philip Resnik. 1995.
\newblock Using information content to evaluate semantic similarity in a
  taxonomy.
\newblock \emph{arXiv preprint cmp-lg/9511007}.

\bibitem[{Rush et~al.(2015)Rush, Chopra, and Weston}]{rush2015neural}
Alexander~M. Rush, Sumit Chopra, and Jason Weston. 2015.
\newblock \href {https://doi.org/10.18653/v1/D15-1044} {A neural attention
  model for abstractive sentence summarization}.
\newblock In \emph{Proceedings of the 2015 Conference on Empirical Methods in
  Natural Language Processing}, pages 379--389, Lisbon, Portugal. Association
  for Computational Linguistics.

\bibitem[{Russo et~al.(2020)Russo, Hollenstein, Musat, and
  Zhang}]{russo2020control}
Giuseppe Russo, Nora Hollenstein, Claudiu~Cristian Musat, and Ce~Zhang. 2020.
\newblock \href {https://doi.org/10.18653/v1/2020.findings-emnlp.33} {Control,
  generate, augment: A scalable framework for multi-attribute text generation}.
\newblock In \emph{Findings of the Association for Computational Linguistics:
  EMNLP 2020}, pages 351--366, Online. Association for Computational
  Linguistics.

\bibitem[{Saito et~al.(2020)Saito, Nishida, Nishida, Otsuka, Asano, Tomita,
  Shindo, and Matsumoto}]{saito2020length}
Itsumi Saito, Kyosuke Nishida, Kosuke Nishida, Atsushi Otsuka, Hisako Asano,
  Junji Tomita, Hiroyuki Shindo, and Yuji Matsumoto. 2020.
\newblock \href {https://arxiv.org/abs/2001.07331} {Length-controllable
  abstractive summarization by guiding with summary prototype}.
\newblock \emph{ArXiv preprint}, abs/2001.07331.

\bibitem[{Schick and Sch{\"u}tze(2021)}]{schick2021s}
Timo Schick and Hinrich Sch{\"u}tze. 2021.
\newblock \href {https://doi.org/10.18653/v1/2021.naacl-main.185} {It{'}s not
  just size that matters: Small language models are also few-shot learners}.
\newblock In \emph{Proceedings of the 2021 Conference of the North American
  Chapter of the Association for Computational Linguistics: Human Language
  Technologies}, pages 2339--2352, Online. Association for Computational
  Linguistics.

\bibitem[{See et~al.(2017)See, Liu, and Manning}]{see2017get}
Abigail See, Peter~J. Liu, and Christopher~D. Manning. 2017.
\newblock \href {https://doi.org/10.18653/v1/P17-1099} {Get to the point:
  Summarization with pointer-generator networks}.
\newblock In \emph{Proceedings of the 55th Annual Meeting of the Association
  for Computational Linguistics (Volume 1: Long Papers)}, pages 1073--1083,
  Vancouver, Canada. Association for Computational Linguistics.

\bibitem[{Shin et~al.(2020)Shin, Razeghi, Logan~IV, Wallace, and
  Singh}]{shin2020eliciting}
Taylor Shin, Yasaman Razeghi, Robert~L Logan~IV, Eric Wallace, and Sameer
  Singh. 2020.
\newblock Eliciting knowledge from language models using automatically
  generated prompts.
\newblock In \emph{Proceedings of the 2020 Conference on Empirical Methods in
  Natural Language Processing (EMNLP)}, pages 4222--4235.

\bibitem[{Song et~al.(2020)Song, Wang, Feng, Liu, and
  Liu}]{song2020controlling}
Kaiqiang Song, Bingqing Wang, Zhe Feng, Ren Liu, and Fei Liu. 2020.
\newblock \href {https://aaai.org/ojs/index.php/AAAI/article/view/6420}
  {Controlling the amount of verbatim copying in abstractive summarization}.
\newblock In \emph{The Thirty-Fourth {AAAI} Conference on Artificial
  Intelligence, {AAAI} 2020, The Thirty-Second Innovative Applications of
  Artificial Intelligence Conference, {IAAI} 2020, The Tenth {AAAI} Symposium
  on Educational Advances in Artificial Intelligence, {EAAI} 2020, New York,
  NY, USA, February 7-12, 2020}, pages 8902--8909. {AAAI} Press.

\bibitem[{Stiennon et~al.(2020)Stiennon, Ouyang, Wu, Ziegler, Lowe, Voss,
  Radford, Amodei, and Christiano}]{stiennon2020learning}
Nisan Stiennon, Long Ouyang, Jeffrey Wu, Daniel~M. Ziegler, Ryan Lowe, Chelsea
  Voss, Alec Radford, Dario Amodei, and Paul~F. Christiano. 2020.
\newblock \href
  {https://proceedings.neurips.cc/paper/2020/hash/1f89885d556929e98d3ef9b86448f951-Abstract.html}
  {Learning to summarize with human feedback}.
\newblock In \emph{Advances in Neural Information Processing Systems 33: Annual
  Conference on Neural Information Processing Systems 2020, NeurIPS 2020,
  December 6-12, 2020, virtual}.

\bibitem[{Tan et~al.(2020)Tan, Qin, Xing, and Hu}]{tan2020summarizing}
Bowen Tan, Lianhui Qin, Eric Xing, and Zhiting Hu. 2020.
\newblock \href {https://doi.org/10.18653/v1/2020.emnlp-main.510} {Summarizing
  text on any aspects: A knowledge-informed weakly-supervised approach}.
\newblock In \emph{Proceedings of the 2020 Conference on Empirical Methods in
  Natural Language Processing (EMNLP)}, pages 6301--6309, Online. Association
  for Computational Linguistics.

\bibitem[{Wolf et~al.(2019)Wolf, Debut, Sanh, Chaumond, Delangue, Moi, Cistac,
  Rault, Louf, Funtowicz et~al.}]{wolf2019huggingface}
Thomas Wolf, Lysandre Debut, Victor Sanh, Julien Chaumond, Clement Delangue,
  Anthony Moi, Pierric Cistac, Tim Rault, R{\'e}mi Louf, Morgan Funtowicz,
  et~al. 2019.
\newblock \href {https://arxiv.org/abs/1910.03771} {Huggingface's transformers:
  State-of-the-art natural language processing}.
\newblock \emph{ArXiv preprint}, abs/1910.03771.

\bibitem[{Xiao et~al.(2022)Xiao, Miculicich, Liu, He, and
  Carenini}]{https://doi.org/10.48550/arxiv.2212.10819}
Wen Xiao, Lesly Miculicich, Yang Liu, Pengcheng He, and Giuseppe Carenini.
  2022.
\newblock \href {https://arxiv.org/abs/2212.10819} {Attend to the right
  context: A plug-and-play module for content-controllable summarization}.

\bibitem[{Zhong et~al.(2022)Zhong, Liu, Ge, Mao, Jiao, Zhang, Xu, Zhu, Zeng,
  and Han}]{zhong2022unsupervised}
Ming Zhong, Yang Liu, Suyu Ge, Yuning Mao, Yizhu Jiao, Xingxing Zhang, Yichong
  Xu, Chenguang Zhu, Michael Zeng, and Jiawei Han. 2022.
\newblock Unsupervised summarization with customized granularities.
\newblock \emph{Findings of EMNLP 2022}.

\bibitem[{Zhong et~al.(2021)Zhong, Yin, Yu, Zaidi, Mutuma, Jha, Awadallah,
  Celikyilmaz, Liu, Qiu, and Radev}]{zhong-etal-2021-qmsum}
Ming Zhong, Da~Yin, Tao Yu, Ahmad Zaidi, Mutethia Mutuma, Rahul Jha,
  Ahmed~Hassan Awadallah, Asli Celikyilmaz, Yang Liu, Xipeng Qiu, and Dragomir
  Radev. 2021.
\newblock \href {https://doi.org/10.18653/v1/2021.naacl-main.472} {{QMS}um: A
  new benchmark for query-based multi-domain meeting summarization}.
\newblock In \emph{Proceedings of the 2021 Conference of the North American
  Chapter of the Association for Computational Linguistics: Human Language
  Technologies}, pages 5905--5921, Online. Association for Computational
  Linguistics.

\end{thebibliography}
\bibliographystyle{acl_natbib}

\appendix
\section{Implementation Details}
We list the implementation details for models. 
\paragraph{HP+SP} 
For HP+SP on MAC-Doc, we load the HP trained model first and set different learning rates for the language model and prefixes, i.e., $3e-5, 1e-6$ separately, and remove the $Len$ prefix from the model. This is because we find that the HP model obtains high performance with $Len$ related attributes very well, due to the pretrained BART-large-CNN checkpoint. Using prefix tuning or tuning the language model with a large learning rate will hurt the performance (Section~\ref{sec:pretrain}). For HP+PE on MAC-Dial, we only set the different learning rates, but we do not load the checkpoint or remove the $Len$ prefix. This is because the CNN pretrained checkpoint is not significantly beneficial for MAC-Dial (Section~\ref{sec:pretrain}).

\paragraph{BART} model is a pretrained BART model which only prepends the hard prompt of topic and speaker to the input, which means it does not control the rest of the attributes. This is the baseline to justify if we control these three attributes.

\section{Annotator Details}
We have four annotators with native English 
background. Before the pilot test, we also supply annotators with professional training for high-quality annotation and provide annotation visualization tools for the annotators to regularize the annotation process. For each sample, we ask the annotators to inspect the quality and decide to keep the annotation or discard it due to difficulty or errors. We combine the annotations of each week to form the \ours~ dataset with careful processing: we discard the invalid samples reported by the annotators and use a program to filter out the other invalid samples with empty or wrong text.

\section{Annotation Guidelines}
We write annotation guidelines of \ours~for two purposes. First, the guidelines are used as our criteria to evaluate annotators during the pilot test. Second, during annotation, we provide annotation guidelines to the annotators and ask them to carefully follow them. For both purposes, the guidelines are a key step to ensure the quality of the whole annotation process. Thus, we pick out some of the details in the guidelines. Note that the following paragraphs are directly copied from the guideline document and shared across all four annotators.

 \paragraph{Speakers annotation criteria.}  A dialogue may contain multiple speakers. if we specify certain speaker names as the control attribute, it means we only care about what these speakers say in the dialogue. So we need to focus on the dialogue turns spoken by these speakers and write the summary for them.

 \paragraph{Topics annotation criteria.} Topic is represented by a set of keywords (usually) copied from the dialogue. A dialogue may contain multiple topics, we need to summarize the content that is only related to the given topic. 

 \paragraph{Length annotation criteria.} \textit{Normal length}: the length of the summary should equal 15\% - 25\% of the related text spans. E.g., the dialogue contains 2000 words, and the related text span for the labeled speaker contains 1000 words. Then we need to write 15\% - 25\% x 1000 = 150 - 250 words for the summary. \textit{Long summary}: the length of the summary should equal 30\%-35\% of the related text spans. \textit{Short summary}: the length of the summary should equal 5\%-10\% of the related text spans. These criteria should be \textit{dynamically modified}, the target of length control is to differentiate the length of the different outputs. We can adjust the criteria a little bit if the lengths of the three types of summaries are too similar.
 
\paragraph{Extractiveness annotation criteria.} \textit{Normal extractiveness}: the same as a natural summary that humans will write. 
\textit{High extractiveness}: copy more sentences/tokens from the source text compared with normal extractiveness.
\textit{Full extractiveness}: copy all the sentences/tokens from the source text.
Again, this can be modified if we can better differentiate summaries with different abstractiveness.

 \paragraph{Specificity annotation criteria.} \textit{Normal specificity}: the same as a natural summary that humans will write.
\textit{High specificity}: include more descriptive content in the source text compared with normal specificity.

\section{Examples of the \ours~Dataset}
Table~\ref{tab:examples} shows five examples of our proposed MAC-Doc dataset. Note that sample 2 and 3, sample 4 and 5 only differs in \textit{Len}, \textit{Ext}, and \textit{Spe}.
 \begin{table*}[!t]
    \centering
    \small
    \begin{tabular}{lp{13cm}}
    \Xhline{5\arrayrulewidth}

        Source text & (CNN)Jackson Gordon is no ordinary 21-year-old. By day he is an industrial design student at Philadelphia University, but Gordon has another side to him -- a side altogether darker, tougher, and more enigmatic. Hanging in his workshop Gordon has a full suit of armor plating, cape, and cowl -- matte black and built to stop a knife. Gordon has an alter ego: the Dark Knight himself, Batman. You might expect his origin story to be cloaked in mystery, but speaking to CNN Gordon is quick to explain how the transformation took place. ... Perhaps because of their versatility and the small matter of copyright issues, those that go on sale will not feature the iconic bat symbol. Gordon says his fledgling business will remain small whilst he's at University -- he has to finish his studies after all, and won't be using the project towards his degree credits. For now the Batsuit and Armatus Design will remain a one man operation: such is the life of a superhero." \\\Xhline{5\arrayrulewidth}
        Attributes  &  Length: short; Extractiveness: normal; Specificity: normal; Topic: (No Topic Specified); \\\midrule
        
        Gold & Jackson Gordon, a 21-year-old industrial design student at Philadelphia University built a Batsuit that is resistant to stabs, knife slashes, and high impacts. According to Gordon, this is a second attempt at building the suit after an earlier attempt five years ago.
        \\
        \midrule
        HP & The transformation of Jackson Gordon, a 21-year-old industrial design student at Philadelphia University, into a Batman fan has happened. Gordon has created a full suit of armor plating, cape and cowl with suede detailing.
         \\\Xhline{5\arrayrulewidth}
         Attributes & Length: normal; Extractiveness: normal; Specificity: normal; Topic: industrial design student;
         \\\midrule
         Gold & Apart from being an industrial design student, Gordon is also a Shaolin Kung Fu expert and has started a business making jackets and cowls but plans to focus on studies first.
         \\\midrule
         HP & The industrial design student Jackson Gordon, 21, is no ordinary student. Gordon has created a replica of the iconic Batman suit with an alter ego named after the Dark Knight.
         \\ \Xhline{5\arrayrulewidth}
         Attributes & Length: normal; Extractiveness: \textbf{full}; Specificity: \textbf{high}; Topic: industrial design student;
         \\\midrule
         Gold & By day he is an industrial design student; Gordon is also an expert in Shaolin Kung Fu; He has already begun manufacturing the cowls for the public.
         \\\midrule
         HP & By day, an industrial design student, Gordon, has another side to him; a side altogether darker, tougher and more enigmatic; Gordon has an alter ego, the Dark Knight himself, Batman; as elaborate as his design was, it lacked the functionality or the authenticity of the genuine article. 
         \\
         \Xhline{5\arrayrulewidth}
        Attributes & Length: normal; Extractiveness: normal; Specificity: normal; Topic: conventional materials;
        \\\midrule
        Gold & Gordon chose unconventional materials to build the Batsuit ensuring that every part was protected whether it had armor plates or not.
        \\\midrule
        HP & In order to avoid using conventional materials, Gordon used memory foam, built around key areas to \"squish and compress\" areas to dissipate the impact of blows, also used Kevlar as the base fabric.
        \\\Xhline{5\arrayrulewidth}
        Attributes & Length: \textbf{long}; Extractiveness: normal; Specificity: normal; Topic: conventional materials;
        \\\midrule
        Gold & Gordon chose unconventional materials, using Kevlar for slash resistance, a form of memory foam for impact absorption, ABS plastic for armor plates, and polyurethane for the cowl.
        \\\midrule
        HP & Eschewing conventional materials, Gordon opted for a form of memory foam, built around key areas to \"squish and compress,\" dissipating the impact of blows; also used Kevlar as the base fabric, making it cut and slash resistant to bladed weapons, but breathable and wearable all day.
        \\

    \Xhline{5\arrayrulewidth}
    \end{tabular}
    \caption{Five case studies on MAC-Doc. 
    }
    \label{tab:examples}
\end{table*}

\end{document}